\documentclass[runningheads]{llncs}
\usepackage{graphicx}
\usepackage{tikz}
\usepackage{comment}
\usepackage{amsmath,amssymb} % define this before the line numbering.
\usepackage{color}
\usepackage{subcaption}

\usepackage[accsupp]{axessibility}  % Improves PDF readability for those with disabilities.

\newcommand{\posea}{\phi}
\newcommand{\poseb}{\rho}
\newcommand{\nerf}{\text{NeRF}}
\newcommand{\embed}{\ell}
\newcommand{\imageshort}{x}
\newcommand{\imagelong}{x}
\newcommand{\remove}[1]{\ignorespaces}
\usepackage{amsmath}
\usepackage{enumitem}

\usepackage{graphicx}
\usepackage{amsmath}
\usepackage{amssymb}
\usepackage{booktabs}
\usepackage{multicol}
\usepackage{algpseudocode}
\usepackage{algorithm}
\usepackage{paralist}
\usepackage{xcolor}
\usepackage{enumitem}
\usepackage{hyperref}

\DeclareMathOperator*{\argmin}{arg\,min}
\begin{document}
% \linenumbers
\pagestyle{headings}
\mainmatter
\def\ECCVSubNumber{4188}  % Insert your submission number here

\title{Neural-Sim: Learning to Generate Training Data with NeRF} % Replace with your title

\begin{comment}
\titlerunning{ECCV-22 submission ID \ECCVSubNumber} 
\authorrunning{ECCV-22 submission ID \ECCVSubNumber} 
\author{Anonymous ECCV submission}
\institute{Paper ID \ECCVSubNumber}
\end{comment}
%******************

% \begin{comment}
\titlerunning{Neural-Sim}

\author{Yunhao Ge\inst{1} \and
Harkirat Behl\inst{2}{$^*$}\and
Jiashu Xu\inst{1}{$^*$} \and
Suriya Gunasekar\inst{2} \and
Neel Joshi \inst{2} \and \\
Yale Song\inst{2} \and
Xin Wang\inst{2} \and
Laurent Itti \inst{1} \and
Vibhav Vineet \inst{2}
}
\authorrunning{Y. Ge et al.}
\institute{University of Southern California \and 
% \email{\{yunhaoge, jiashuxu, briannlz, itti\}@usc.edu}
Microsoft Research
% \email{Vibhav.Vineet@microsoft.com}\\
}

% \end{comment}
%******************
\maketitle

\begin{abstract}
    %%%%%%%%% ABSTRACT
% \begin{abstract}
% Neural Radiance Field (NeRF) has become a powerful method for generating novel views of complex scenes and is widely used in various graphics and vision tasks. However, how does NeRF perform on substituting the traditional graphics pipeline and generating useful data to help downstream tasks (e.g., object detection and segmentation) remain underexplored. To fill this gap, we focus on using NeRF, as a generative model, to overfit any object and synthesize required images that are demanded for training an object detector (discriminative model). To learn an automatic data generation that can automatically find the optimal NeRF generation parameters (object pose, zoom in/out factor) based on the change of validation scenario (data), we propose Bilevel NeRF-Synthesis Optimization (BiNSO) which optimize both the neural rendering model and downstream detection model in an end-to-end pipeline. Qualitative and quantitative experiments demonstrate that (1) NeRF model can substitute the traditional graphics neural rendering pipeline (e.g., BlenderProc) and synthesize useful images to help downstream tasks. (2) Our optimization pipeline can automatically find the optimal NeRF generation parameter which can synthesis useful data for downstream task training.
% Traditional approaches for 
Training computer vision models usually requires collecting and labeling vast amounts of imagery under a diverse set of scene configurations and properties. This process is incredibly time-consuming, and it is challenging to ensure that the captured data distribution maps well to the target domain of an application scenario. Recently, synthetic data has emerged as a way to address both of these issues. However, existing approaches either require human experts to manually tune each scene property or use automatic methods that provide little to no control; this requires rendering large amounts of random data variations, which is slow and is often suboptimal for the target domain.
We present the first fully differentiable synthetic data  pipeline that uses Neural Radiance Fields (NeRFs) in a closed-loop with a target application's loss function.  Our approach generates data on-demand, with no human labor, to maximize accuracy for a target task.
We illustrate the effectiveness of our method on synthetic and real-world object detection tasks.
%, demonstrating its ability to generate data on-demand for different scenarios and domains.
We also introduce a new ``YCB-in-the-Wild'' dataset and benchmark that provides a test scenario for object detection with varied poses in real-world environments. Code and data could be found at \textcolor{magenta}{\url{https://github.com/gyhandy/Neural-Sim-NeRF}}.

% \end{abstract}
% \dots

\end{abstract}
\def\thefootnote{*}\footnotetext{Equal contribution as second author}\def\thefootnote{\arabic{footnote}}
    \section{Introduction}
    \label{sec:intro}
    %%%%%%%%% Introduction
% \vspace{-10pt}
% \section{Introduction}

%%%%%%
%%%%%%   Teaser Image (Image-1)
%%%%%%
\begin{figure}[t]
\begin{center}
\includegraphics[width=\linewidth]{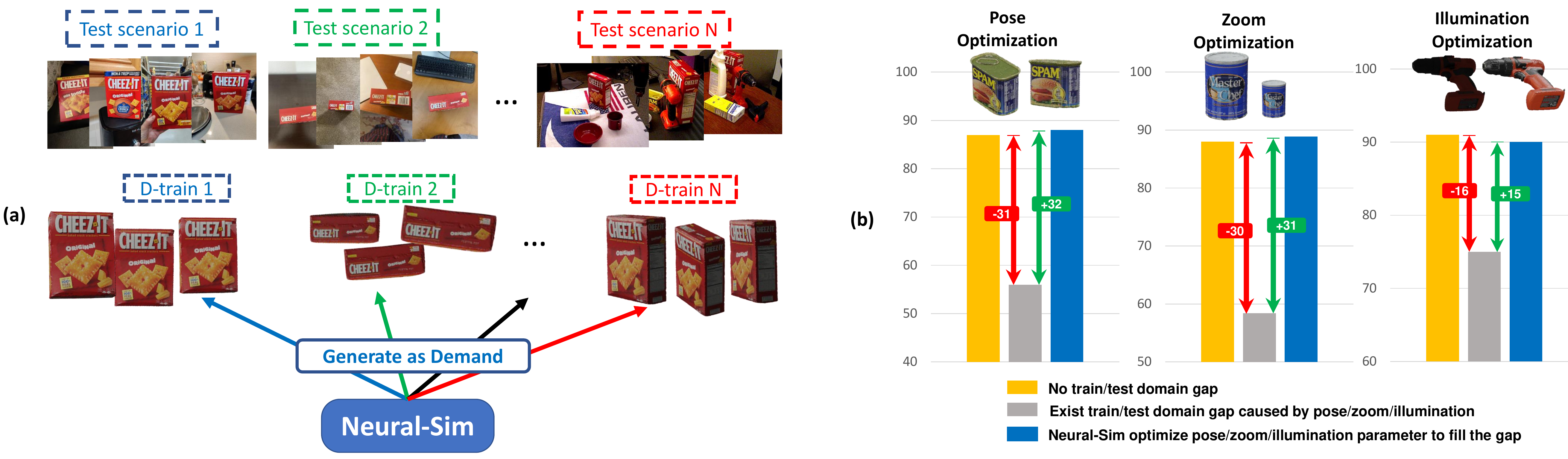}
\end{center}
   \caption{(a) On-demand synthetic data generation:  Given a target task and a test dataset, our approach ``Neural-sim" generates data on-demand using a fully differentiable synthetic data generation pipeline which maximises accuracy for the target task. (b) Train/test domain gap causes significant detection accuracy drop (yellow bar to gray bar). We dynamically optimize the render parameters (pose/zoom/illumination) to generate the best data to fill the gap (blue bar). }
\label{fig1}
\end{figure}

%%%%%%%%%%%%%%%%%%%%%%%%%%%%%%%%%%%%%%%%%%
%%% Traditional pipeline and challenges %%
%%%%%%%%%%%%%%%%%%%%%%%%%%%%%%%%%%%%%%%%%%
The traditional pipeline for building computer vision models involves collecting and labelling vast amounts of data, training models with different configurations, and deploying it to test environments \cite{he2017mask,ren2017faster,Shelhamer_2017}. 
Key to achieving good performance is collecting training data that mimics the test environment with similar properties relating to the object (pose, geometry, appearance), camera (pose and angle), and scene (illumination, semantic structures)\cite{objectnet2019}. 

However, the traditional pipeline does not work very well in many real-world applications as collecting large amounts of training data which captures all variations of objects and environments is quite challenging.
Furthermore, in many applications, users may want to learn models for unique objects with novel structures, textures, or other such properties. 
Such scenarios are very common particularly in business scenarios where there is desire to create object detectors for new products introduced in the market.

%%%%%%%%%%%%%%%%%%%%%%%%%%%%%%%%%%%%%%%%%%
%%% Synthetic data for Training %%
%%%%%%%%%%%%%%%%%%%%%%%%%%%%%%%%%%%%%%%%%%
Recent advances in rendering, such as photo-realistic renderers \cite{denninger2019blenderproc,georgiev2018arnold} and generative models (GANs \cite{brock2018large}, VAEs \cite{VAEKingma,betavae}), have brought the promise of generating high-quality images of complex scenes. 
This has motivated the field to explore synthetic data as source of training data \cite{doersch2019sim2real,IlgMSKDB_cvpr17,dwibedi2017cut,HandaPBSC_corr15a,hodan2019photorealistic,RichterHK_iccv17,RosSMVL_cvpr16,TremblayTB_corr18,ge2020zero,ZhangSYSLJF_cvpr17,ge2022dall}.
However, doing so in an offline fashion has similar issues as the traditional pipeline. While it alleviates certain difficulties, e.g., capturing camera/lighting variations, it create dependency on 3D asset creation, which is time-consuming.
%For high performance, synthetic data should be automatically generated by learning the optimal scene properties for the downstream task.

%%%%%%%%%%%%%%%%%%%%%%%%%%%%%%%%%%%%%%%%%%
%%%%%%%%%%%%%%%%% NeRF %%%%%%%%%%%% 
%%%%%%%%%%%%%%%%%%%%%%%%%%%%%%%%%%%%%%%%%%
%\textbf{Why NeRF } (i) high quality images (ii) differentiability (iii) control over scene parameters (iv) less human intervention
Recently, a new image generation technique called the Neural Radiance Field (NeRF) \cite{mildenhall2020nerf} was introduced as a way to replace the traditional rasterization and ray-tracing graphics pipelines with a neural-network based renderer.  This approach can generate high-quality novel views of scenes without requiring explicit 3D understanding. More recent advancements in NeRFs allow to control other rendering parameters, like illumination, material, albedo, appearance, etc. \cite{nerv2021,martin2021nerf,zhang2021nerfactor,bi2020neural,jang2021codenerf}.
As a result, they have attracted significant attention and have been widely adopted in various graphics and vision tasks \cite{gafni2021dynamic,bi2020neural,nerv2021,park2021nerfies}. 
NeRF and their variants possess some alluring properties: (i) differentiable rendering,  (ii) control over scene properties unlike GANs and VAEs, and (iii) they are data-driven in contrast to traditional renderers which require carefully crafting 3D models and scenes. 
These properties make them suitable for generating the optimal data on-demand for a given target task.

%%%%%%%%%%%%%%%%%%%%%%%%%%%%%%%%%%%%%%%%%%
%%%%%%%%%%%%%%%%% Our Method %%%%%%%%%%%% 
%%%%%%%%%%%%%%%%%%%%%%%%%%%%%%%%%%%%%%%%%%
To this end, we propose a bilevel optimization process to jointly optimize neural rendering parameters for data generation and model training. 
% Our proposed method leverages the benefits of the differentiability of data generation through NeRF. 
%
Further, we also propose a reparameterization trick, sample approximation, and patch-wise optimization methods for developing a memory efficient optimization algorithm.

To demonstrate the efficacy of the proposed algorithm, we evaluate the algorithm on three settings: controlled settings in simulation, on the YCB-video dataset \cite{xiang2017posecnn}, and in controlled settings on YCB objects captured in the wild. This third setting is with our newly created ``YCB-in-the-wild'' dataset, which involves capturing YCB objects in real environments with control over object pose and scale. Finally, we also provide results showing the interpretability of the method in achieving high performance on downstream tasks. Our key contributions are as follows:

%   {\bf Overall Contribution:} 
  (1) To the best of our knowledge, for the first time, we show that NeRF can substitute the traditional graphics pipeline and synthesize useful images to train downstream tasks (object detection). %\vspace{-7pt}

%   {\bf Technical Contribution:} 
  (2) We propose a novel bilevel optimization algorithm to automatically optimize rendering parameters (pose, zoom, illumination) to generate 
  optimal data for downstream tasks using NeRF and its variants. %\vspace{-7pt}

%   {\bf Experimental Contribution:}
  (3) We demonstrate the performance of our approach on controlled settings in simulation, controlled settings in YCB-in-wild and YCB-video datasets. We release YCB-in-wild dataset for future research. %\vspace{-5pt}

% \begin{itemize}
% % \vspace{-7pt}
% %\begin{itemize}[leftmargin=0.5cm]\setlength{\itemsep}{-1pt}\vspace{-7pt}
%   \item 
% %   {\bf Overall Contribution:} 
%   To the best of our knowledge, for the first time, we show that NeRF can substitute the traditional graphics pipeline and synthesize useful images to train downstream tasks (object detection). %\vspace{-7pt}
%   \item 
% %   {\bf Technical Contribution:} 
%   We propose a novel bilevel optimization algorithm to automatically optimize rendering parameters (pose, zoom, illumination) to generate 
%   optimal data for downstream tasks using NeRF and its variants. %\vspace{-7pt}
%   \item 
% %   {\bf Experimental Contribution:}
%   We demonstrate the performance of our approach on controlled settings in simulation, controlled settings in YCB-in-wild and YCB-video datasets. We will release YCB-in-wild dataset for future research. %\vspace{-5pt}
% \end{itemize}

    \section{Related work}
    \label{sec:related_work}
    % \vspace{-5pt}
% \section{Related Work}
% \vspace{-5pt}

%\TODO{Needs some updates.}
%% 1
% Prior work learns non-differentiable simulator parameters. Ours learn differentiable simulator parameters.

%%%% Auto-Simulate, learning to simulate, they are competitor  
%%%% meta-sim, meta-sim2, 
%%%% Orthogonal to our work.

%% 2
% GAN or any other differentiable simulator
%%% Generic approach and NeRF has shown to

% \vspace{-5pt}
% \subsection{Neural Rendering Methods }
% \vspace{-5pt}
\textbf{Traditional Graphics rendering methods} can synthesize high-quality images with controllable image properties, such as object pose, geometry, texture, camera parameters, and illumination \cite{RichterHK_iccv17,denninger2019blenderproc,georgiev2018arnold,hodan2019photorealistic,richter2016playing}.
Interestingly, NeRF has some important benefits over the traditional graphics pipelines, which make it more suitable for learning to generate synthetic datasets. 
First, NeRF learns to generate data from new views based only on image data and camera pose information. 
In contrast, the traditional graphics pipeline requires 3D models of objects as input.
Getting accurate 3D models with correct geometry, material, and texture properties generally requires human experts (i.e. an artist or modeler). This, in turn, limits the scalability of the traditional graphics pipeline in large-scale rendering for many new objects or scenes.
%In contrast, NeRF does not suffer from these issues as it is a data-driven approach, needs only images and corresponding camera pose information to generate high quality images from new  view points.
Second, NeRF is a differentiable renderer, thus allowing backpropagation through the rendering pipeline for learning how to control data generation in a model and scene-centric way.
%so learning data generation through NeRF involves accurate gradient through data generation component.
% Structure-from-motion pipeline may only generate noisy version of the 3D models (some examples are shown in Fig. \ref{}).
%However they requires human expert manually handcrafting each property which are expensive and slow, they are also non-differentiable methods makes rendering parameters learning harder. 

\noindent{\bf Deep generative models}, such as GANs \cite{goodfellow2014generative,brock2018large}, VAEs \cite{VAEKingma,betavae} and normalizing flows \cite{NormalizingFlow} are differentiable and require less human involvement.
However, most of them do not provide direct control of rendering parameters. 
While some recent GAN approaches allow some control 
\cite{VIGAN,SteerableGAN,ge2020pose} over parameters, it is not as explicit and can mostly only change the 2D properties of images. 
Further, most generative models need a relatively large dataset to train. 
%Neural Radiance Field (NeRF) and its update works [] are more and more powerful and widely used in various graphics and vision tasks[]. 
In comparison, NeRF can generate parameter-controllable high-quality images and requires a lesser number of images to train. 
Moreover, advancements in NeRF now allow the control of illumination, materials, and object shape alongside camera pose and scale \cite{nerv2021,martin2021nerf,zhang2021nerfactor,bi2020neural,jang2021codenerf}. We use NeRF and their variants (NeRF-in-the-wild \cite{martin2021nerf}) to optimize pose, zoom and illumination as representative rendering parameters.

%We use NeRF as our Neural-Sim generator and automatically learn the best rendering parameters.

% \vspace{-15pt}
% \subsection{Learning simulator parameters }
% \vspace{-5pt}
\noindent{\bf Learning simulator parameters.}
Related works in this space focus on learning non-differentiable simulator parameters for e.g., learning-to-simulate (LTS) \cite{ruiz2018learning}, Meta-Sim \cite{kar2019metasim}, Meta-Sim2 \cite{devaranjan2020meta}, Auto-Sim \cite{behl2020autosimulate}, and others \cite{yang2019learning,Ganin2018SynthesizingPF,louppe2017adversarial}. Our work in contrast has two differences: (i) a difference in the renderer used (NeRF vs traditional rendering engines), and (ii) a difference in the optimization approach.
We discuss the different renderers and their suitability for this task in the previous subsection.
% However, it is important to note that our work is orthogonal to the above mentioned works \cite{ruiz2018learning,kar2019metasim,behl2020autosimulate,yang2019learning}. This is because just like NeRF is an alternate to traditional renderers, Neural-sim is an alternate to Meta-sim and Auto-sim.  %\neel{I'm not sure why is it orthogonal}

LTS \cite{ruiz2018learning} proposed a bilevel optimization algorithm to learn simulator parameters that maximized accuracy on downstream tasks. It assumed both data-generation and model-training as a black-box optimization process and used REINFORCE-based \cite{Williams_1992} gradient estimation to optimize parameters. This requires many intermediate data generation steps. Meta-sim \cite{kar2019metasim} is also a REINFORCE based
approach, which requires a grammar of
scene graphs. Our approach does not use scene grammar.
Most similar to our work is the work of Auto-Simulate \cite{behl2020autosimulate} that proposed a local approximation of the bilevel optimization to efficiently solve the problem.
However, since they optimized non-differentiable simulators like Blender \cite{denninger2019blenderproc} and Arnold \cite{georgiev2018arnold}, they used REINFORCE-based \cite{Williams_1992} gradient update. %\TODO{Add reference to variance of gradient is large.} 
Further, they have not shown optimization of pose parameter whose search space is very large. In comparison, our proposed Neural-Sim approach can learn to optimize over pose parameters as well.
%
%
%NeRF based neural renderer as a data generation component. This allows us to get accurate gradients through the data generation pipeline. 

%%%%%%%%% BODY TEXT
% \section{Method: Neural-Sim}

    %\input{latex/method_complete_before_SGedits}
    % \input{method2.tex}
    % \input{latex/method_old_har}

    % \section{Method}
    \label{sec:method}
    % \vspace{-15pt}
\section{Neural-Sim}
% \vspace{-10pt}
The goal of our method is to automatically synthesize optimal training data to maximize
accuracy for a target task. In this work, we consider object detection as our target task.
% Neural networks require extensive amounts of training data. 
Furthermore, in recent times, NeRF and its variants (NeRFs) have been used to synthesize high-resolution photorealistic images for complex scenes \cite{nerv2021,martin2021nerf,zhang2021nerfactor,bi2020neural,jang2021codenerf}.
% Recently, a new image generation technique called the Neural Radiance Field (NeRF) \cite{mildenhall2020nerf} was introduced, which promises to synthesize high-resolution photorealistic images for complex scenes.
This motivates us to explore NeRFs as potential sources of generating training data for computer vision models. 
We propose a technique to optimize rendering parameters of NeRFs to generate the \textit{optimal set} of images for training object detection models. 
%The goal of our approach is to generate optimal synthetic data that maximizes accuracy on the target task. 
%
%Over the last year, NeRF and its variants \cite{mildenhall2020nerf,nerv2021,martin2021nerf,park2021nerfies,barron2021mip} have been shown to generate photo-realistic images for complex scenes while controlling the rendering process through scene properties like camera pose, zoom, and lighting. In this work,  we investigate if NeRF could be programmed to not just generate any data, but rather generate an \textit{optimal set} of images for training neural network models for a target task. 

\textbf{NeRF model:} 
NeRF \cite{mildenhall2020nerf,lin2020nerfpytorch} takes as input the viewing direction (or camera pose) denoted as $V=(\posea,\poseb)$, and renders an image $\imagelong=\nerf(V)$ of a scene as viewed along $V$. %to optimize for camera pose as the rendering parameter. 
Note that our proposed technique is broadly applicable to differentiable renderers in general. In this work, we also optimize NeRF-in-the-wild (NeRF-w) \cite{martin2021nerf} as it allows for appearance and illumination variations alongside pose variation. We first discuss our framework for optimizing the original NeRF model and later we discuss optimization of NeRF-w in Section \ref{nerfw}.

%\vspace{-1em}
\textbf{Synthetic training data generation:} 
Consider a parametric probability distribution $p_\psi$ over rendering parameters $V$, where $\psi$ denotes the parameters of the distribution. It should be noted that $\psi$ corresponds to all rendering parameters including pose/zoom/illumination, here, for simplicity, we consider $\psi$ to denote pose variable.
To generate the synthetic training data, we first sample  rendering parameters $V_{1}, V_2,..., V_{N}\sim p_{\psi}$. We then use NeRF to generate synthetic training images $\imageshort_i = \nerf(V_i)$ with respective rendering parameters $V_i$. We use an off-the-shelf foreground extractor to obtain labels $y_1,y_2,\ldots,y_N$. the training dataset thus generated is denoted as $D_{train}=\{(\imageshort_1,y_1),(\imageshort_2,y_2),\ldots,(\imageshort_N,y_N)\}$. 

%%%%%%
%%%%%%   Fig-pipeline
%%%%%%
\begin{figure*}[t!]
% \vspace{-10pt}
\begin{center}
\includegraphics[width=\linewidth]{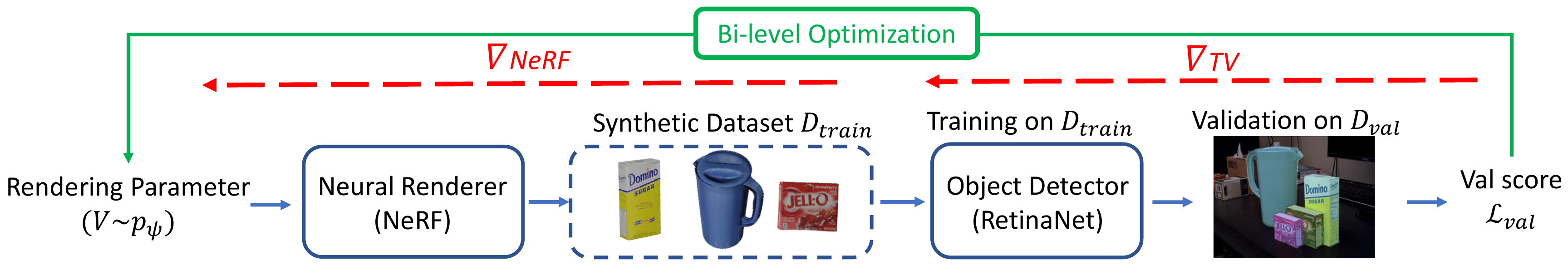}
\end{center}
% \vspace{-4mm}
   \caption{Neural-Sim pipeline: Our pipeline finds the optimal parameters for generating views from a trained neural renderer (NeRF) to use as training data for object detection.  The objective is to find the optimal NeRF rendering parameters $\psi$ that can generate synthetic training data $D_{train}$, such that the model (RetinaNet, in our experiments) trained on $D_{train}$, maximizes accuracy on a downstream task represented by the validation set $D_{val}$.}
\label{fig:pipeline}
% \vspace{-15pt}
\end{figure*}

%\vspace{-1em}
\textbf{Optimizing synthetic data generation} 
Our goal is to optimize over the rendering distribution $p_\psi$ such that training an object detection model on $D_{train}$ leads to good performance on $D_{val}$. 
We formulate this problem as a bi-level optimization \cite{colson2007overview,franceschi2018bilevel,behl2020autosimulate} as below:
%%%%%%%%%%%%
%%%%%%%%%%%% Eq.1 bilevel optimization
%%%%%%%%%%%%
\begin{subequations}
\begin{align}
\underset{\psi}{\min}  \ \ \mathcal{L}_{val}(\hat{\theta}(\psi));  \ \ \ \ \ \
s.t.  \ \ \hat{\theta}(\psi) \in \underset{\theta} \argmin  \ \  \mathcal{L}_{train}(\theta, \psi),\label{eq:inner_problem}
\end{align}\label{eq:original_bilevel_problem}
% \vspace{-10pt}
\end{subequations}
% \begin{subequations}
% \begin{align}
% \underset{\psi}{\min} & \ \ \mathcal{L}_{val}(\hat{\theta}(\psi))  \\
% s.t.  & \ \ \hat{\theta}(\psi) \in \underset{\theta} \argmin  \ \  \mathcal{L}_{train}(\theta, \psi),\label{eq:inner_problem}
% \end{align}\label{eq:original_bilevel_problem}
% \end{subequations}

\noindent where $\theta$ denotes the parameters of the object detection model, $\mathcal{L}_{train}(\theta, \psi)=\mathbb{E}_{V\sim p_{\psi}} l(\imageshort,\theta)\approx\frac{1}{N}\sum_{i=1}^N l({\imageshort_i,\theta})$ is the training loss over the synthetic dataset from NeRF,\footnote{For simplicity, we have dropped the dependence of loss $\ell$ on labels $y$} and $\mathcal{L}_{val}$ is the  loss on the task-specific validation set $D_{val}$. %Observe that we now have a optimization over two sets of parameters: object detection parameters $\theta$ and NeRF rendering parameters $\Psi$. 

The bi-level optimization problem in \eqref{eq:original_bilevel_problem} is challenging to solve;  for example, any gradient based algorithm would need access to an efficient approximation of $\nabla_{\psi}\hat{\theta}(\psi)$, which in turn requires propagating gradients through the entire training trajectory of a neural network. Thus, we look to numerical approximations to solve this problem.  Recently, Behl~et.~al.~\cite{behl2020autosimulate} developed a technique for numerical gradient computation based on local approximation of the bi-level optimization. 
Without going into their derivation, we borrow the gradient term for the outer update, which at time step $t$ takes the form: 
% \vspace{-5pt} 
%%%%%%%%%%%%
%%%%%%%%%%%% Eq.8 approximation dLval/d\psi
%%%%%%%%%%%%
{%\small
\begin{align}
\frac{\partial\mathcal{L}_{val}(\hat{\theta}(\psi))}{\partial\psi}\Bigg\lvert_{\psi = \psi_t} \!\!\!\!\!\!\!\!\!\!\!\approx & 
-\overbrace{\frac{\partial}{\partial\psi} \Big[\frac{\partial \mathcal{L}_{train}\big(\hat{\theta}(\psi_t),\psi) }{\partial\theta}\Big]^{T}}^{\nabla_{NeRF}} \Bigg\lvert_{\psi = \psi_t}%\nonumber\\
%& 
\underbrace{\mathcal{H}(\hat{\theta}(\psi_{t}), \psi)^{-1}
\frac{d\mathcal{L}_{val}(\hat{\theta}(\psi_t))}{d\theta}}_{\nabla_{TV}}.
\label{eq:2}
\end{align}}
% \vspace{-10pt}

We have divided the gradient term into two parts: $\nabla_{NeRF}$  corresponds to backpropagation through the dataset generation from NeRF, and $\nabla_{TV}$ corresponds to approximate backpropagation through training and validation (Fig.~\ref{fig:pipeline}). 
 $\nabla_{TV}$ is computed using the conjugate gradient method \cite{behl2020autosimulate}. However, \cite{behl2020autosimulate} treated the data generation as a black box and used REINFORCE \cite{williams1992simple} to compute the approximate gradient because they used non-differentiable renderers for data generation. However, REINFORCE is considered noisy process and is known to lead to high-variance estimates of gradients. In contrast, NeRF is differentiable, which gives us tools to obtain more accurate gradients. We propose an efficient technique for computing $\nabla_{NeRF}$, which we discuss in the next section.

\subsection{Backprop through data generation from NeRF}

A good gradient estimation should possess the following properties: (i) high accuracy and low noise, (ii) computational efficiency, (iii) low memory footprint. %Note that this eliminates the method of finite differences because it is noisy and computationally inefficient.
We leverage different properties of NeRF, i.e., its differentiability and pixel-wise rendering, to design a customized technique which satisfies the above properties. %Note that the following description, easily adapts to any neural renderer that provides control over other scene rendering parameters like object pose, object scale, lighting, etc. 

%Recall data generation from NeRF described earlier and depicted in Figure~\ref{fig:pipeline}. 
\remove{
Data is generated from NeRF as follows:
$\psi$ denotes the parameter of the distribution over camera pose parameters, and starting from this distribution, we sample poses as $V_j = (\phi_j$, $\poseb_j)$, which are then used to render images with NeRF. 
Note that the following description also applies to cases where $\psi$ corresponds to rendering parameters like object pose, object scale, etc.

Prior work treated data generation as a black box and used Reinforce \cite{williams1992simple} to compute the gradient. This was because they used non-differentiable renderers for data generation. 
However, the gradients computed using Reinforce are only approximate and noisy.
In contrast, NeRF is differentiable, which allows us to get more accurate gradients.

A good gradient estimation should possess the following desirable properties: (i) computational efficiency, (ii) high accuracy and low noise, (iii) low memory footprint.
Note that this eliminates the method of finite differences because it is noisy and computationally inefficient.
In this work, we leverage NeRF's characteristics to design an estimation technique which satisfies the above properties. 

}
%%%%%% Everything beloe is from new version
In computation of $\nabla_{NeRF}$ in \eqref{eq:2}, we approximate $\mathcal{L}_{train}(\theta, \psi)$ using samples in $D_{train}$ as $\mathcal{L}_{train}(\theta, \psi)\approx\frac{1}{N}\sum_{i=1}^N l({x_i,\theta})$. Using chain rule we then have partial derivative computation over $l({x,\theta})$ as follows: 
 %$\mathcal{L}_{train}(\theta, \Psi)=\mathbb{E}_{\psi\sim p_{\Psi}} l_y(\imageshort,\theta)\approx\frac{1}{N}\sum_{i=1}^N\l_i({\imageshort_i,\theta})$
%%%%%%%%%%%%
%%%%%%%%%%%% Eq.9 approximation E() chain role
%%%%%%%%%%%%
{\begin{align}
\frac{\partial}{\partial\psi}\bigg[\frac{\partial l(x_i, \hat{\theta}(\psi_{t}))}{\partial\theta}\bigg]\!=\! \bigg[\frac{\partial(\frac{\partial l(x_i, \hat{\theta}(\psi_{t}))}{\partial\theta})}{\partial x_i}\bigg]
\!\bigg[\frac{\partial x_i}{\partial V_i}\bigg]\!\bigg[\frac{d V_i}{d\psi}\bigg]\label{eq:9}
\end{align}}
The first term is the second order derivative through an object detection network and can be computed analytically for each image $x_{i}$. The second term is the gradient of the rendered image w.r.t NeRF inputs, which again is well defined and can be obtained by backpropagating through the differentiable NeRF rendering $x_i = \nerf(V_i)$. While both these terms have exact analytical expressions, naively computing and using them in \eqref{eq:2} becomes impractical even for small problems (see below in Tool2 and Tool3 for details and proposed solutions). 
%We present efficient solutions to overcome this issue in Section \ref{twice} and \ref{patch_wise} .
Finally the third term $\frac{dV_i}{d\psi}$ requires gradient computation over probabilistic sampling $V_i\sim p_{\psi}$. We consider $p_\psi$ over discretized bins of pose parameters. For such discrete distributions $\frac{dV_i}{d\psi}$ is not well defined. Instead, we approximate this term using a reparameterization technique described below in Tool1. We summarize our technical tools below:
 \begin{compactitem}
 \item  For distributions $p_\psi$ over a discrete bins of pose parameters, we propose a reparametrization of $\psi$ that provides efficient approximation of $\frac{dV_i}{d\psi}$  (Tool1).
 \item We dramatically reduce memory and computation overhead of implementing the gradient approximation in \eqref{eq:2} using a new \textit{twice-forward-once-backward} approach (Tool2). Without this new technique the implementation would require high computation involving large matrices and computational graphs. 
 \item Even with the above technique, the computation of first and second terms in \eqref{eq:9} has a large overhead in terms of GPU memory that depends on image size. We overcome this using a patch-wise gradient computation approach described in Tool 3.
 \end{compactitem}

\subsubsection{Tool 1: Reparametrization of pose sampling}\label{reparametrization} 
NeRF renders images $x_j$ using camera pose $V_j$=($\posea_i$, $\poseb_j$), where $\posea_j \in [0, 360], \poseb_j \in [0, 360]$. For simplicity we describe our method for optimizing over just $\posea$, while keeping $\poseb$ fixed to be uniform. %We can essentially duplicate this process to optimize over $\poseb$ (e.g., experiments in Section~\ref{sec:5.4}). 

We discretize the pose space into $k$ equal sized bins over the range of $\phi$ as $B_1=\big[0,\frac{360}{k}\big),B_2=\big[\frac{360}{k},\frac{360\times 2}{k}\big),\ldots$.
and define the distribution over $\posea$ as the categorical distribution with $p_i$ as the probability of $\phi$ belonging to $B_i$. This distribution is thus parametrized by $\psi \equiv p=[p_1, ..., p_k]$. 

To back propagate through the sampling process, we approximate the sample from the categorical distribution by using Gumble-softmax “reparameterization trick” with parameters $y\in\mathbb{R}^k$, where $y_i$ are given as follows:
%%%%%%%%%%%%
%%%%%%%%%%%% Eq.11 Gumble distribution
%%%%%%%%%%%%
{\begin{align}
% \vspace{-10pt}
y_i  &= \text{GS}_i(p) = \exp[(G_i+log(p_i)) /  \tau] / \sum_{j} \exp[(G_i+log(p_j)) / \tau],
% \vspace{-10pt}
\label{eq:gumbel}
% \vspace{-50pt}
\end{align}}
% {\begin{align}
% y_i  &= \text{GS}_i(p) = \frac{\exp(\frac{G_i+log(p_i)}{ \tau})}{\sum_{j} \exp(\frac{G_i+log(p_j)}{\tau})},\label{eq:gumbel}
% \end{align}}
where $G_i \sim Gumbel(0,1)$ are i.i.d. samples from the standard Gumbel distribution and $\tau$ is temperature parameter. The random vector $y$ defined as above satisfies the property that the coordinate (index) of the largest element in $y\in\mathbb{R}^k$ follows the categorical distribution with parameter $p$.%, i.e., if $y_i$ is the largest dimension, then bin $B_i$ is the selected bin. 

\begin{figure*}[tb]
\begin{center}
\includegraphics[width=\linewidth]{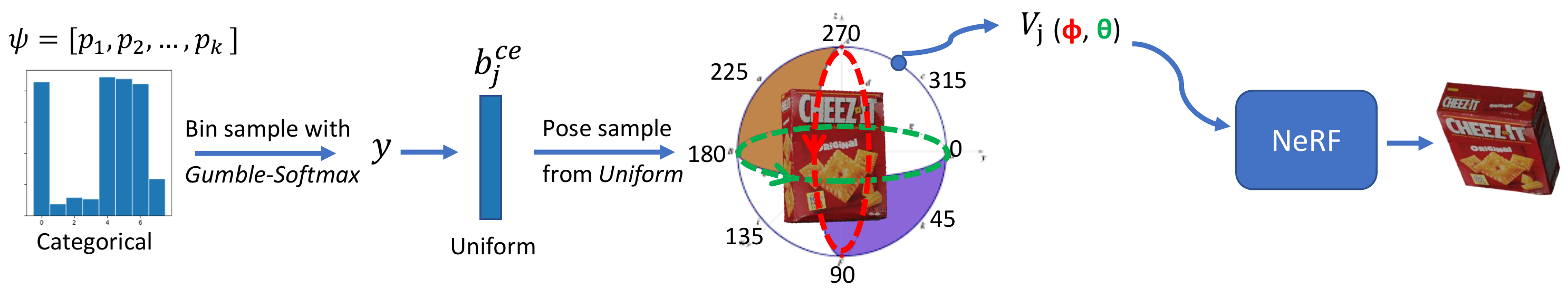}
\end{center}
% \vspace{-15pt}
   \caption{A concrete example to one time sample, starting form a particular value of $\psi$, we can follow reparametrization sampling and obtain a pose.  Each sample represents a pose that is input in NeRF to render one image.}
\label{fig:example}
% \vspace{-10pt}
\end{figure*}

We now approximate sampling from the categorical distribution (see Figures \ref{fig:gumble} and \ref{fig:example} for depiction). 
Denote the bin center of $B_i$ as $\bar{B}^{ce}_{i}={360(i-0.5)}/{k}$; and the bin range as $\bar{b}^{ra}={360}/{k}$. We generate $V_j=(\posea_j,\poseb_j)\sim p_{\psi}$ as below:
\begin{compactitem}
\item Generate $y_i$'s for $i=1,2,\ldots k$ from \eqref{eq:gumbel}
\item Define $b_{j}^{ce}=\sum_i y_i \bar{B}_i^{ce}$ as the approximate bin center.
\item Define the bin for the $j^\text{th}$ sample centered around $b_{j}^{ce}$ as $[b_{j}^{st},b_{j}^{en}]=[b_{j}^{ce}-\bar{b}^{ra}/2,b_{j}^{ce}+\bar{b}^{ra}/2]$
\item We sample $\phi_j$ from uniform distribution over $[b_{j}^{st},b_{j}^{en}]$ which has a standard reparametrization for diffentiability: $\mathcal{U}(b_{j}^{st},b_{j}^{en})\equiv 
(1-\epsilon) b_{j}^{st}+\epsilon b_{j}^{en}\text{ s.t. }\epsilon\sim\mathcal{U}(0,1)$. 
\item $\poseb_j\sim\mathcal{U}[0,360]$, or can follow same process as $\posea_j$.
\end{compactitem}

Note that in general the approximate bin centers $b_{j}^{ce}$ need not be aligned with original categorical distribution, however we can control the approximation using the temperature parameter $\tau$. As $\tau \rightarrow 0$, $y$ will be a one-hot vector and exactly emulate sampling from categorical distribution. 

%%%%%%
%%%%%%   Fig-3
%%%%%%
% \begin{figure}[tb]
% \begin{center}
% \includegraphics[width=0.5\linewidth]{Fig/Fig-gumble.pdf}
% \end{center}
% \vspace{-3mm}
%   \caption{Bin sampling:  We first discretize the pose space into a set of $k$ bins, which we will then sample to generate the view parameters for the NeRF.  To backpropagate through the sampling process, we approximate the sample from the categorical (i.e. bin) distribution by using a Gumble-softmax “reparameterization trick”.  Within each bin we sample uniformly.}
% \label{fig:gumble}
% \vspace{-5pt}
% \end{figure}

% \vspace{5pt}
%%%%%%
%%%%%%   Fig-3
%%%%%%
\begin{minipage}{0.54\linewidth}
    % \centering
    \includegraphics[width=2.4in]{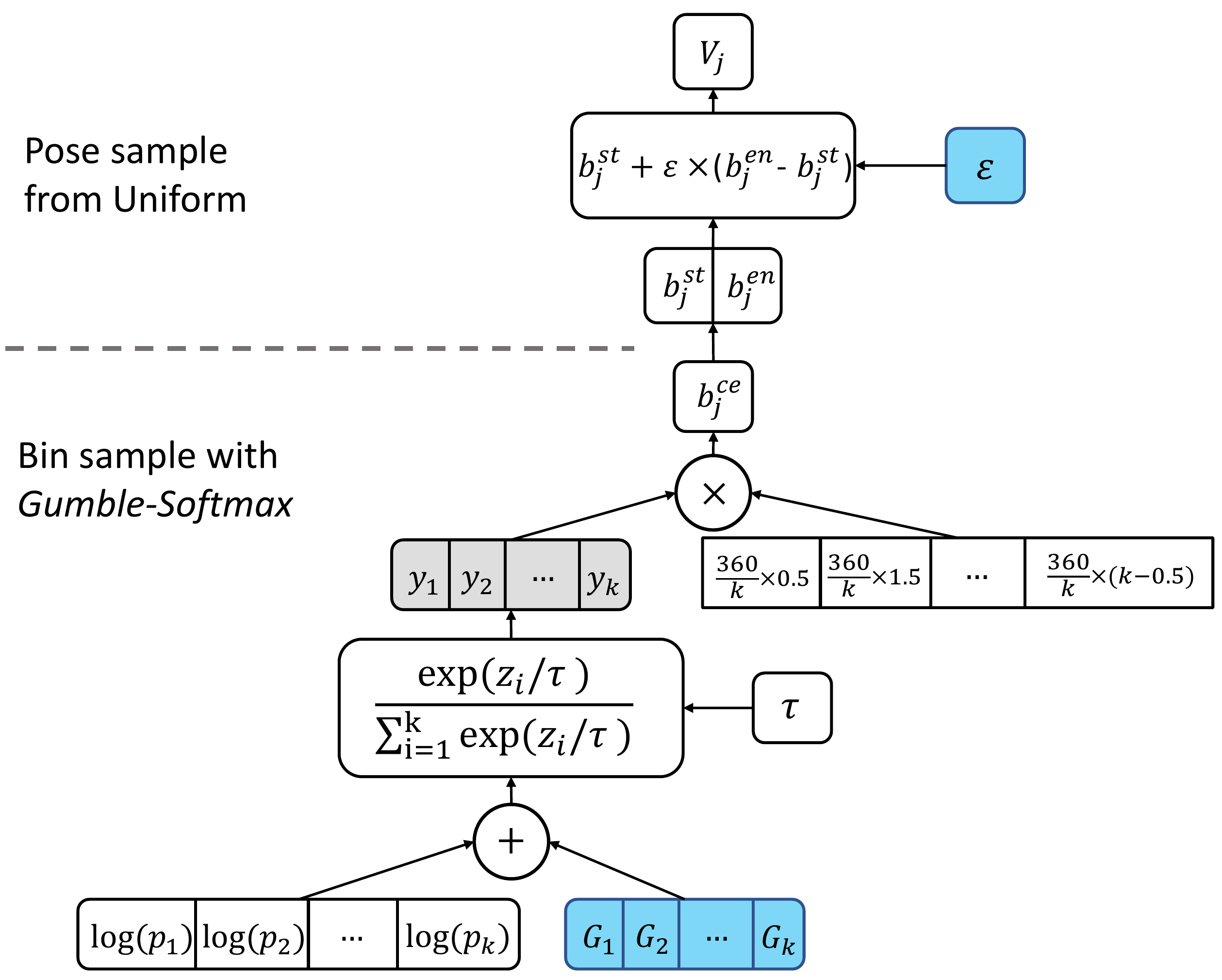}
\end{minipage}
\begin{minipage}{0.40\linewidth}
    \centering
    % \small
    \captionof{figure}{Bin sampling:  We first discretize the pose space into a set of $k$ bins, which we will then sample to generate the view parameters for the NeRF.  To backpropagate through the sampling process, we approximate the sample from the categorical (i.e. bin) distribution by using a Gumble-softmax “reparameterization trick”.  Within each bin we sample uniformly.}
    % \znote{please verify my edits}}
\label{fig:gumble}
\end{minipage}

%%%%%%
%%%%%%   Fig-example
%%%%%%
We now have the full expression for approximate gradient of $\nabla_{NeRF}$ using \eqref{eq:9} and reparametrization as follows:
\begin{equation}
% \vspace{-3pt}
\begin{aligned}
\nabla_{NeRF}\!
%&\frac{\partial}{\partial\psi} \underset{x \in D_{train}}{\mathbb{E}} \bigg[\frac{\partial l(x, \hat{\theta}(\psi_{t}))}{\partial\theta}\bigg]
%= \frac{1}{N} \sum\limits_{j=1}^N  \frac{\partial(\frac{\partial l(x_j, \hat{\theta}(\psi_{t}))}{\partial\theta})}{\partial x_j} \frac{\partial x_j}{\partial\psi} \\
%&= \frac{1}{N} \sum\limits_{j=1}^N 
%\frac{\partial(\frac{\partial l(x_j, \hat{\theta}(\psi_{t}))}{\partial\theta})}{\partial x_j} 
%\frac{\partial x_j}{\partial V_j} \frac{\partial V_j}{\partial U (bi_{bottom},bi_{upper})}\\
%&\frac{\partial U (bi_{bottom},bi_{upper})}{\partial y} \frac{\partial y}{\partial \psi} \\
\approx\!
\frac{1}{N}\! \sum\limits_{j=1}^N\! 
\frac{\partial(\!\frac{\partial l(x_j, \hat{\theta}(\psi_{t}))}{\partial\theta}\!)}{\partial x_j} 
\frac{\partial x_j}{\partial V_j} \frac{\partial V_j}{\partial (b_j^{st},b_j^{en})}\frac{\partial (b_i^{st},b_i^{en})}{\partial y} \frac{\partial y}{\partial p} .
\label{eq:chain}
\end{aligned} 
% \vspace{-5pt}
\end{equation}
Below we present tools that drastically improve the compute and memory efficiency and are crucial for our pipeline.

\subsubsection{Tool 2: Twice-forward-once-backward} \label{twice}
The full gradient update of our bi-level optimization problem involves using the approximation of $\nabla_{NeRF}$ in \eqref{eq:chain} and back in \eqref{eq:2}. This computation has three terms with the following dimensions: 

\noindent 
\begin{inparaenum}[(1)]
\item $\frac{\partial(\frac{\partial l(x_j, \hat{\theta}(\psi_{t}))}{\partial\theta})}{\partial x_j} \in \mathbb{R}^{m\times d}$,
\item $\frac{\partial x_j}{\partial\psi} \in \mathbb{R}^{d\times k}$, 
\item $\nabla_{TV}=\mathcal{H}(\hat{\theta}(\psi_{t}), \psi)^{-1}\frac{d\mathcal{L}_{val}(\hat{\theta}(\psi_t))}{d\theta}\in \mathbb{R}^{m\times 1}$,
\end{inparaenum}
where $m=|\theta|$ is the $\#$ of parameters in object detection model, $d$ is the $\#$ of pixels in $x$, and $k$ is $\#$ of pose bins. 

Implementing eq. \eqref{eq:2} with the naive sequence of $(1)$-$(2)$-$(3)$ involves computing and multiplying large matrices of sizes  $m\times d$ and $d\times k$. Further, this sequence also generates a huge computation graph. These would lead to prohibitive memory and compute requirements as $m$ is often in many millions. On the other hand, if we could follow the sequence of $(3)$-$(1)$-$(2)$, then we can use the produce of $1\times m$ output of $(3)$ to do a weighted autograd which leads  computing and storing only vectors rather than matrices.  However, the computation of $(3)$ needs the rendered image involving forward pass of $(2)$  (more details in supp.).

To take advantage of the efficient sequence, we  propose a twice-forward-once backward method where we do two forward passes over NeRF rendering. In the first forward path, we do not compute the gradients, we only render images to form $D_{train}$ and save random samples of $y,\phi_j$ used for rendering. We then compute $(3)$ by turning on gradients. In the second pass through NeRF, we keep the same samples and this time compute the gradient $(1)$ and $(2)$. %More details are in supplementary.

\subsubsection{Tool 3: Patch-wise gradient computation}\label{patch_wise}

Even though we have optimized the computation dependence on $m=|\theta|$ with the tool described above,  computing $(1)$-$(2)$ sequence in the above description still scales with the size of images $d$. This too can lead to large memory footprint for even moderate size images (e.g., even with the twice-forward-once-backward approach, the pipeline over a $32\times 32$ image already does not fit into a $2080Ti$ GPU). To optimize the memory further, we propose patch-wise computation, where we divide the image into $S$ patches $x=(x^1,x^2,\ldots,x^S))$ and compute \eqref{eq:9} as follows:
%%%%%%%%%%%%
%%%%%%%%%%%% Eq.15 Memory efficient Patch-wise Optimization
%%%%%%%%%%%%
\begin{equation}
% \vspace{-5pt}
\begin{aligned}
\frac{\partial}{\partial\psi} \frac{\partial l(x, \hat{\theta}(\psi_{t}))}{\partial\theta}
= \sum\limits_{c=1}^{S} 
\frac{\partial(\frac{\partial l(x^{c}, \hat{\theta}(\psi_{t}))}{\partial\theta})}{\partial x^{c}}
\frac{\partial x^{c}}{\partial\psi}.
\end{aligned}
% \vspace{-5pt}
\end{equation}
% \begin{equation}
% \begin{aligned}
% \small
% &\frac{\partial}{\partial\psi} \underset{x \in D_{train}}{\mathbb{E}} \bigg[\frac{\partial l(x, \hat{\theta}(\psi_{t}))}{\partial\theta}\bigg]
% = \frac{1}{N} \sum\limits_{j=1}^N 
% \frac{\partial(\frac{\partial l(x_j, \hat{\theta}(\psi_{t}))}{\partial\theta})}{\partial x_j}
% \frac{\partial x_j}{\partial\psi} \\ 
% &= \frac{1}{S*N} \sum\limits_{j=1}^{S*N} 
% \frac{\partial(\frac{\partial l(x_j^{c}, \hat{\theta}(\psi_{t}))}{\partial\theta})}{\partial x_j^{c}}
% \frac{\partial x_j^{c}}{\partial\psi} 
% \end{aligned}
% \end{equation}
Since NeRF renders an image pixel by pixel, it is easy to compute the gradient of patch with respect to $\psi$ in the memory efficient patch-wise optimization. 
% \vspace{-10pt}
\subsection{Nerf-in-the-wild}
% \vspace{-5pt}
\label{nerfw}
NeRF-in-the-wild (NeRF-w) extends the vanilla NeRF model to allow image dependent appearance and illumination variations such that photometric discrepancies between images can be modeled explicitly. NeRF-w takes as input an appearance embedding denoted as $\embed$ alongside the viewing direction $V$ to render an image as $\imagelong=\nerf(V, \embed)$. For NERF-w, the optimization of pose (V) remains the same as discussed above. 
For efficient optimization of lighting we exploit a noteworthy property of NeRF-w: it allows smooth interpolations between color and lighting. This enables us to optimize lighting as a continuous variable, where the lighting ($\ell$) can be written as an affine function of the available lighting embeddings ($\ell_i$) as $\ell = \sum_i \psi_i * \ell_i$ where $\sum_i \psi_i =1$. To calculate the gradient from Eq. \ref{eq:9}, $\frac{\partial x_i}{\partial \ell}$ is computed in the same way as described above utilizing our tools 2 and 3, and the term $\frac{d \ell}{d\psi}$ is straightforward and is optimized with projected gradient descent.

%To achieve variable relighting and photometric postprocessing, NeRF-w adopt the approach of Genarative Latent Optimization (GLO) \cite{bojanowski2017optimizing} in which each image is assigned a corresponding real-valued appearance embedding vector $l^(a)$ of length $n^(a)$. Using these appearance embeddings as input to only the branch of the network that emits color grants NeRF-w the freedom to vary the emitted radiance of the scene in a particular image while still guaranteeing that the 3D geometry. By setting $n^(a)$ to a small value (we use 48), NeRF-w encourage optimization to identify a continuous space in which illumination conditions can be embedded, thereby enabling smooth interpolations between different color and illumination while geometry is fixed.

    % \section{Experiments}
    \label{sec:experiment}
    % section3
%%%%%%%%%%%
%%%%%%%%%%%
% Experiments
%%%%%%%%%%%
%%%%%%%%%%%
% \vspace{-10pt}
\section{Experiments}
% \vspace{-15pt}
We now evaluate the effectiveness of our proposed Neural-Sim approach in generating optimal training data on object detection task. 
We provide results under two variations of our Neural-Sim method. In the first case, we use Neural-Sim without using bi-level optimization steps. In this case, data from NeRF are always generated from the same initial distribution. The second case involves our complete Neural-Sim pipeline with bi-level optimization updates (Eq. \ref{eq:2}). In the following sections, we use terms NS and NSO for Neural-Sim without and Neural-Sim with bi-level optimization respectively.

%
%versions of our method. In first case, we use Neural-Sim to generate data without any optimization. Here data is generated from uniform 
%
%benefit of using Ne
%
%
%
%on three different scenarios.
%
%obtain the optimal rendering parameter for NeRF to maximize the accuracy on downstream tasks with NeRF synthesized data.
%
%Our proposed Neural-Sim approach use NeRF as a differentiable rendering simulator to synthesize data which can be used to train downstream tasks. 
%
%Neural-Sim also use bilevel optimization to obtain the optimal rendering parameter for NeRF to maximize the accuracy on downstream tasks with NeRF synthesized data. 
%
We first demonstrate that NeRF can successfully generate data for downstream tasks as a substitute for a traditional graphic pipeline (e.g., BlenderProc) (Sec.~\ref{sec:5.1}) with similar performance. Then we conduct experiments to demonstrate the efficacy of Neural-Sim in three different scenarios: controllable synthetic tasks on YCB-synthetic dataset (Sec.~\ref{sec:5.2}); controllable real-world tasks on YCB-in-the-wild dataset (Sec.~\ref{sec:5.3}); general real-world tasks on YCB-Video dataset (Sec.~\ref{sec:5.4}). We also show the interpretable properties of the Neural-Sim approach (NSO) during training data synthesis (Sec.~\ref{sec:5.5}). All three datasets are based on the objects from the YCB-video dataset \cite{xiang2017posecnn,hodan2018bop,calli2015benchmarking}. 
% which is a subset of YCB dataset\cite{calli2015benchmarking}
It contains 21 objects from daily life and provides high-resolution RGBD images with ground truth annotation for object bounding boxes. 
The dataset consists of both digital and physical objects, which we use to create both real and synthetic datasets.

% Implementation
\noindent{\bf Implementation details:}  
%1) \noindent{\bf NeRF}
We train one NeRF-w model for each YCB object using 100 images with different camera pose and zoom factors using BlenderProc.
%
% To train NeRF model for one YCB object, we use BlenderProc to synthesize 100 images with different pose and zoom factor. 
%We explore the training parameter influence on NeRF rendering, such as number and pose range of images used to train NeRF, near and far. 
% some properties of NeRF training and rendering properties such as
%  quality, 
% important implementation details, how many images we use to train nerf, detectron parameter.
% To simplify the optimization, we sample $\theta_j \sim \mathcal{U}[0, 360] $ uniformly, while we focus on the optimization of sampling parameters of $\phi_j$.
%2) \noindent{\bf Object Detector}
We use RetinaNet \cite{lin2017focal} as our downstream object detector. 
To accelerate the optimization, we fix the backbone during training. 
During bi-level optimization steps, we use Gumble-softmax temperature $\tau=0.1$. In each optimization iteration, we render 50 images for each object class and train RetinaNet for two epochs. More details are in the supplementary material.  

\noindent{\bf Baselines:} We compare our proposed approach against two popular state-of-the-art approaches that learn simulator parameters. The first baseline is Learning to simulate \cite{ruiz2018learning} which proposed a REINFORCE-based approach to optimize simulator parameters. Also note that the meta-sim \cite{kar2019metasim} is a REINFORCE-based approach. Next, we consider Auto-Sim \cite{behl2020autosimulate} which proposed an efficient optimization method to learn simulator parameters. We implemented our own version of Learning to simulate work and we received code from the authors of Auto-Sim.

%\VV{Let us give description of dataset here.}
%

% \VV{In the main paper or in the supplementary material, we should add images of all the YCB objects.}

% \section{Implementation}

% \noindent{\bf Implementation details:}  
% 1) \noindent{\bf NeRF}
% To train a NeRF model for one YCB object, we use BlenderProc to synthesize 100 images with different pose and zoom factor. We explore the training parameter influence on NeRF rendering, such as number and pose range of images used to train NeRF, near and far. 
% % some properties of NeRF training and rendering properties such as
% %  quality, 
% % important implementation details, how many images we use to train nerf, detectron parameter.
% % To simplify the optimization, we sample $\theta_j \sim \mathcal{U}[0, 360] $ uniformly, while we focus on the optimization of sampling parameters of $\phi_j$.
% 2) \noindent{\bf Object Detector} We use RetinaNet \cite{lin2017focal} as our downstream object detector. To accelerate the optimization, we fix the backbone during training. 3) \noindent{\bf Bi-level Optimization} we use Gumble-softmax temperature $\tau$ =0.1; In each optimization iteration, we render 50 images for each object class and train RetinaNet 2 epochs. More details are in supplement material.  

% \vspace{-15pt}
\subsection{NeRF to generate data for downstream tasks}
% \vspace{-10pt}
\label{sec:5.1}
First, it is important to show that NeRF is a suitable replacement for a traditional renderer like BlenderProc \cite{denninger2019blenderproc} when generating data for object detection. 
%
%
%In order to create YCB-synthetics dataset, 
To test this, we use YCB-video dataset objects and
%parameter $\psi$ to form $D_{train}^{NeRF}$. 
%
we render images from NeRF and BlenderProc \cite{denninger2019blenderproc} using the same camera pose and zoom parameters.
%
%We also use BlenderProc to synthesize same 4 YCB dataset object images under rendering parameter $\psi$ to form $D_{train}^{Blender}$. 
%
%$\psi$ represents camera pose, zoom in factor and illumination \VV{How are we controlling illumination?}. %
%
We use these images to conduct object detection
% single-object and multi-object detection 
tasks under same training and test setting.
% (more details in supplementary). 
%using both  $D_{train}^{NeRF}$ and $D_{train}^{Blender}$ under same training and test setting (more details in supplementary). 
%
Both object detectors trained on NeRF synthesized images and BlenderProc images have nearly same accuracy. (More details in Supplementary). 
% (90\% mAP on YCB synthetic object detection task). 
% The quantitative results in Fig.~\ref{fig:7} show that NeRF generated data can achieve the same accuracy as that of BlenderProc on downstream object detection tasks.

% %%%%%%
% %%%%%%   Fig-7
% %%%%%%
% \begin{figure}[t!]
% \vspace{-10pt}
% \begin{center}
% % \includegraphics[width=\linewidth]{Fig/Fig-4.png}
% \includegraphics[width=\linewidth]{Fig/Fig-nerfvsblender.pdf}
% \end{center}
% \vspace{-4mm}
%   \caption{Object detection performance using NeRF and Blender synthesized images.}
% \vspace{-10pt}
% \label{fig:7}
% \end{figure}

% \item Controllable dataset. Model performance drop when there were distribution shift. Geometric gap (pose variation, zoom factor variation, illumination )

% \vspace{-10pt}
\subsection{YCB-synthetic dataset}
% \vspace{-5pt}
%\subsection{Neural-Sim solve distribution gap caused performance drop in YCB-synthetic}
\label{sec:5.2}

%%%%%%
%%%%%%   Fig-control, fill gap
%%%%%%
\begin{figure}[t]
\begin{center}
\includegraphics[width=\linewidth]{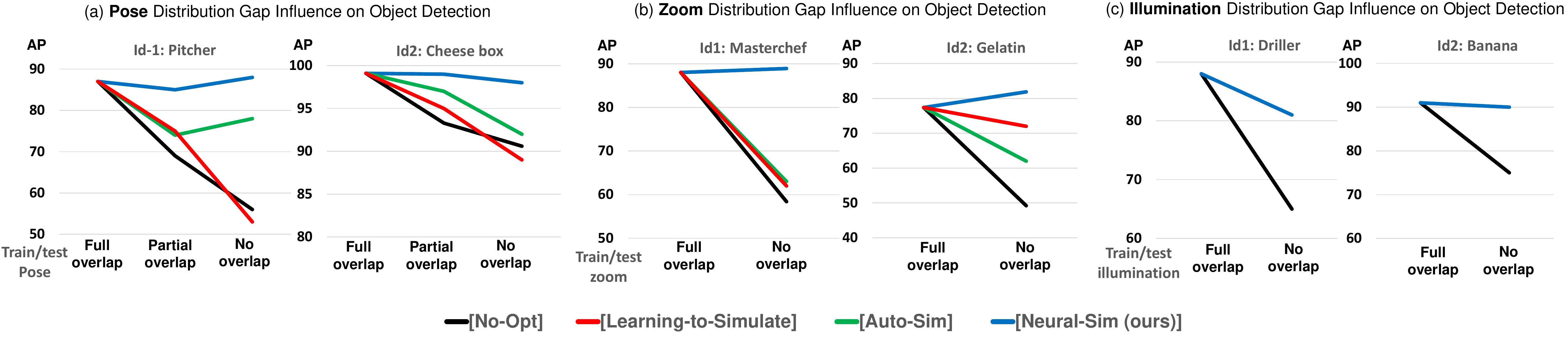}
\end{center}
% \vspace{-4mm}
   \caption{ Neural-Sim performance on YCB-Synthetic. When there are distribution gap between train and test sets ((a) pose (b) zoom (c) illumination gap), with the gap increase, object detection faces larger accuracy drop (black line). With the help of Neural-Sim (NSO) in blue line, the performance drop are filled. Observe improvement of NSO over LTS \cite{ruiz2018learning} (red line) and Auto-Sim \cite{behl2020autosimulate} (green line).
   %     pose               partial             No overlap
   % L2S: pitcher           75                    53
   % L2S: Driller           95                    89
   %     zoom
   % L2S: masterchef         62,  
   % L2S: Gelatin            72, 
   % L2S: mug                75
   %
   % illumination. driller improve from 65 to 81
   % illumimation. banana improve from 75 to 90
   %
   %
%   \neel{fonts need to be bigger in the plots}
   }
% \vspace{-15pt}
\label{fig:5}
\end{figure}

Next, we conduct a series of experiments on a YCB-synthetic dataset to show how NSO helps to solve a drop in performance due to distribution shifts between the training and test data. 
%
%%The distribution gap have been created by changing object pose, zoom factor and illumination difference between training and test sets, which are important parameters in $\psi$. 

%
%In this section, we use Neural-Sim to solve the problem and fill the performance drop by synthesizing useful training images $D_{train}$. 
%
%where test set is synthetically generated using BlenderProc. 
%
%
%Fig.~\ref{fig:2} demonstrate the model performance drop when there were distribution shift between training and testing sets. \VV{Provide quantitative numbers here.}

\noindent{\bf Dataset setting} We select six objects that are easily confused with each other: {\em masterchef} and {\em pitcher} are both blue cylinders and {\em cheezit}, {\em gelatin}, {\em mug} and {\em driller} are all red colored objects. 
To conduct controlled experiments, we generate data with a gap in the distribution of poses between the training and test sets. For this, we divide the object pose space into $k$= 8 bins.
%and define a categorical distribution over the object pose bins (same as Eq.~\ref{eq:bin}). % Do we need to include this.
%. where the bin number $k$= 8 based on the range of pose $\phi$ (same as Eq.~\ref{eq:bin}). 
%To form the object pose distribution gap, we define a categorical distribution over object pose where the bin number $k$= 8 based on the range of pose $\phi$ (same as Eq.~\ref{eq:bin}). 
%
For each objects $o_j$ and pose bin $i$ combination, we use BlenderProc \footnote{BlenderProc is a popular code-base to generate photo realistic synthetic data using traditional graphics pipeline.} to synthesize 100 images.
%\neel{maybe need to say why is groundtruth synthesized with BlenderProc?}
%$o_j$ images with poses belonging to bin $i$. 
%
These images of the six selected objects with pose bin-labels form YCB-synthetic data.
% (\VV{Can we use pose bin-labels rather than pose meta-labels? Or we can define meta-labels.}) % (\VV{Do we need to put YCB-synthetics in mathematics form?}) . 

%
\noindent{\bf Train/test biasness} We create controlled experiments by varying the degree of pose distribution overlap between the training and test sets.
%
%We use the same experimental setup as used in Fig.~\ref{fig:2} where x axis represents the degree of pose distribution overlap between the training and test sets.
%
For each object (e.g. {\em pitcher}) we fix its pose distribution in the test set (e.g. images are generated with pose from bin 1) and change its pose distribution in training set in three ways. First, images are generated with pose with same distribution as test set (bin1 is dominant), uniform distribution (pose values uniformly selected from bin1 to bin 8) and totally different distribution from the test set (other bins are dominant except bin 1). We introduce such pose biasness in two of the six objects, {\em pitcher} and {\em driller}. For other four objects, test images are generated from an uniform distribution. The test set has 600 images (100 images per object).

%In the 6 class object detection task, test set has 600 images (100 image per class). 
%
%For each interest object (e.g. pitcher) we fix its pose distribution in test set (e.g. all images poses are in bin 1) and change its pose distribution in training set: same distribution as test (bin1 are dominant), uniform distribution (pose randomly choose from bin1 to bin 8), totally different distribution as test (other bins dominant except bin 1). 
%
%The change of training pose distribution cause gap between training and test distribution. 
%
%The uninterested classes has no distribution gap which all has uniform distribution. \VV{Let us use slightly better wording for the previous sentence.}

%

\noindent{\bf Results} Quantitative results are shown in Fig.~\ref{fig:5}. 
First, we show the performance of our NS 
%
%NeRF 
based training images rendered using three initial distributions described earlier. 
%
%For the training set, if we use NeRF to synthesize images follow the same distribution setting (fix $\psi$) as the x axis, the performance shows the mAP of the interested class (pitcher or driller) that cause pose distribution gap (full line in Fig.~\ref{fig:6}). 
%
We observe that the object detection performance drops by almost $30 \%$ and $10 \%$ for {\em pitcher} and {\em driller} objects respectively when there is object pose gap between training and test distributions. 
%Note that in this situation, NeRF images are generated without optimization over the pose-distribution.

%
%
Next we show that our Neural-Sim with bi-level optimization (NSO) is able to automatically find the optimal pose distribution of the test set. NeRF then uses the optimal distribution to synthesize training data. The object detection model trained on the optimal data helps  improve performance significantly; average precision accuracy for the {\em pticher} and {\em driller} objects have been improved by almost $30\%$ and $10\%$, respectively. The blue lines in Fig.~\ref{fig:5} represent the performance of 
NSO
%
%Neural-Sim approach after optimization 
%
which fill the gap caused by distribution mismatch. Note there is similar significant improvement in experiments where there is gap in camera zoom when using the proposed NSO approach.

We compare our NSO with the recent work Learning-to-simulate (LTS) \cite{ruiz2018learning} and Auto-Sim \cite{behl2020autosimulate} that use REINFORCE for non-differentiable simulator optimization (Fig.~\ref{fig:5}(a)(b)). We observe that on pose optimization, the proposed NSO achieves almost  $34\%$ improvement over LTS and $11\%$ improvement over Auto-Sim on on the pitcher object. On zoom optimization, NSO achieves almost $27\%$ improvement over LTS and $26\%$ improvement over Auto-Sim on Masterchef object. This highlights the gradients from differentiable NSO are more effective and can generate better data than REINFORCE based LTS and Auto-Sim.

\noindent {\bf Experiments on illumination optimization.} To verify the effectiveness of Neural-Sim on illumination, we substitute vanilla NeRF model with NeRF-w. We conduct similar experiments as the pose and zoom experiments in Sec.~\ref{sec:5.2} on illumination with YCB-synthetic dataset. The results show in Fig.~\ref{fig:5}(c). NSO has great performance on illumination optimization with $16\%$ and $15\%$ improvements on driller and banana objects respectively.

\noindent {\bf Large scale YCB-Synthetic dataset experiments}
Here we highlight the results of our large-scale experiments on the YCB-synthetic dataset. Experiments demonstrate that our proposed NSO approach helps to solve a drop in performance due to distribution shifts between the train and test sets.
We use the same setting as previous experiment except we conduct object detection on all 21 objects on the YCB-Synthetic dataset. 
We create controlled experiments by varying the degree of pose distribution overlap between the training and test sets. For each object, we fix its pose distribution in the test set and change its pose distribution in the training set: training images are generated from totally different distributions from the test set. The test set has 2100 images (100 images per object). The experiment results are shown in Table.~\ref{table:ycb_syn_21}. We compare the proposed NS and NSO approaches with the baseline Auto-Sim \cite{behl2020autosimulate} method. Note that our proposed NSO achieves improvements of almost 14 $\%$ and 13 $\%$ points over NS and Auto-Sim baselines respectively.

% We observe that the proposed NSO approach achieves very large improvement over the baselin Auto-Sim method in many experiment scenarios. For example, on {\em pitcher} object, starting from uniform and random initial distributions, we observe an improvement of almost 40\% and 60\% improvement respectively. Further, we also observe improvement on {\em cheeze box} and multi-modal experiments . These experiments demonstrate effectiveness of the proposed NSO method on finding correct parameters over the baseline Auto-Sim approach.

%%%%%%%%% 
%%%%%%%%% table-1
%%%%%%%%% 
\begin{table}[t]
%   \small
%   \footnotesize
  \scriptsize
  \caption{Large scale YCB-synthetic experiments}
  \label{table:ycb_syn_21}
  \centering
\begin{tabular}{c |c  c  c  c  c  c  c  c  c  c c}
    \toprule
Objects & \bf{mAP} &\begin{tabular}{@{}l@{}}master \\ chef can\end{tabular} & \begin{tabular}{@{}l@{}}cracker \\ box\end{tabular} & \begin{tabular}{@{}l@{}}sugar \\ box\end{tabular} & \begin{tabular}{@{}l@{}} tomato \\ soup can\end{tabular} & \begin{tabular}{@{}l@{}} mustard\\ bottle\end{tabular} & \begin{tabular}{@{}l@{}} tuna fish\\ can \end{tabular} & \begin{tabular}{@{}l@{}} pudding\\box \end{tabular} & \begin{tabular}{@{}l@{}} gelatin\\ box\end{tabular} & \begin{tabular}{@{}l@{}} potted\\ meat can\end{tabular} & banana \\
    \midrule
 NS        & 68.4 &93.5 & 96.6 & 58.3 & 83.9 & 78.4 & 44.3 & 78.0 & 65.2 & 55.3 & 89.4 \\
 Auto-Sim  & 69.3 &96.0 & 82.5 & 92.3 & 37.4 & 81.3 & 52.0 & 80.6 & \bf{79.4} & 74.4 & 83.4 \\
 NSO       & \bf{82.1} &\bf{98.5} & \bf{98.4} & \bf{98.2} & \bf{81.8} & \bf{90.5} & \bf{64.6} & \bf{84.1} & 57.6 & \bf{92.2} & \bf{91.6}  \\
    \midrule
%  & bleach cleanser & bowl & mug & power drill & wood block & scissor & large marker & large clamp & extra large clamp & foam brick & mAP \\
Objects & \begin{tabular}{@{}l@{}} pitcher\\ base\end{tabular} & \begin{tabular}{@{}l@{}}bleach \\ cleanser\end{tabular} & bowl & mug & \begin{tabular}{@{}l@{}} power \\ drill\end{tabular} & \begin{tabular}{@{}l@{}} wood\\ block\end{tabular} & scissor & \begin{tabular}{@{}l@{}} large\\marker \end{tabular} & \begin{tabular}{@{}l@{}} large\\ clamp\end{tabular} & \begin{tabular}{@{}l@{}} extra large\\ clamp\end{tabular} & \begin{tabular}{@{}l@{}} foam\\ brick\end{tabular} \\
    \midrule
 NS        & 29.0 & 49.9 & 78.7 & 46.8 & 89.3 & 97.8 & \bf{67.9} & 42.9 & 47.8 & 72.7 & \bf{69.6}  \\
 Auto-Sim  & 7.7 & 81.5 & 78.3 & 60.0 & 83.2 & 95.6 & 64.1 & 41.5 & 46.6 & \bf{79.0} & 57.9  \\
 NSO       & \bf{83.5} & \bf{93.4} & \bf{98.5} & \bf{87.9} & \bf{93.6} & \bf{98.7} & 55.3 & \bf{56.9} & \bf{50.8} & 78.6 & 68.2 \\
     \bottomrule
\end{tabular}
\end{table}

% effectiveness of our proposed differentiable optimization NSO method in generating optimal data for the downstream object detection task over REINFORCE based Auto-Sim approach.
%Neural-Sim optimization.
%\neel{It's a little confusing that there is Neural-Sim and Neural-Sim optimization.  Maybe we should call them two more different things?  Or use and acronym: NS and NSO?}

%However, our proposed Neural-Sim approach is able to automatically find the optimal pose distribution of the test set that has been used to synthesize the training data $D_{train}$. The object detection model trained on the optimal data $D_{train}$ helps to significantly improve the performance and accuracy is improved by almost $30\%$ and $10\%$ on the pitcher and driller objects. The dashed lines represent the performance of Neural-Sim approach which fill the gap caused by distribution mismatch.

%Neural-Sim can find the optimal object pose distribution in rendering parameter $\psi$ to synthesize on demand $D_{train}$, which can maximize the validation performance after training a RetinaNet with $D_{train}$. Take the same training distribution as starting point, Neural-Sim can keep optimize the object pose $\psi$ and synthesize on demand data. The dash line represent the performance of Neural-Sim which fill the gap caused by distribution mismatch.
%

% % figure
% same plot but show improvements. before optimization and after optimization.

% \vspace{-10pt}
%%%%%%
%%%%%%
%%%%%% YCB-in-the-wild dataset
%%%%%%
%%%%%%
\subsection{YCB-in-the-wild dataset}
\label{sec:5.3}
% \vspace{-1.3mm}

To evaluate the performance of the proposed NS and NSO approaches %Neural-Sim 
%
%\neel{again here do you mean the whole thing or the version with or without opimization} 
%
on a real world dataset, we have created a real world {\em YCB-in-the-wild} dataset. The dataset has 6 YCB objects in it,
%in the {\em YCB-in-the-wild} dataset 
which are same as in the {\em YCB-synthetic} dataset: {\em masterchef, cheezit, gelatin, pitcher, mug} and {\em driller}. All images are captured using a smartphone camera in a common indoor environments: living room, kitchen, bedroom and bathroom, under natural pose and illumination. We manually labelled each image with object bounding boxes. Further, to explore the effect of distribution shifts on the object detection task, we manually labelled the object pose in each image using the the same eight bins discussed earlier. 
%
%
%
%We use the same categorical distribution setting as Sec.~\ref{sec:5.2}, where all pose sphere are separated into eight bins. 
%
The dataset consists of total around  1300 test images with each object having over 200 images.
%
%
%Each object class has around 200 test images and all dataset has over 1300 test images. 
Some of the images from the dataset are shown in the Fig \ref{fig1}. 
% and Fig.~\ref{fig:vis}. 
% \VV{Add reference.}
%
%Thus we have manually labelled object detection information and object pose meta-labels. 
We will release the original images, ground truth object detection labels and pose bin-labels.
%for future research. 
% \VV{Let us go over the above description again.}
%
%the data synthesis of Neural-Sim, $YCB-wild$ has controllable pose information, where we meta labelled the object pose of each images. 
%
%We use the same categorical distribution setting as Sec.~\ref{sec:5.2}, where all pose sphere are separated into 8 bins. Each class object in $YCB-wild$ has over 200 images and all dataset has over 1300 images.

To explore the performance of the NS and NSO
%Neural-Sim method 
under the training and test distribution gap on the {\em YCB-in-the-wild}, we use the same experiment setup as in Sec.~\ref{sec:5.2}. The test images are selected from {\em YCB-in-the-wild} and training images are synthesized by NeRF. The training data is generated under two categorical distributions: uniform distribution and a random bin as dominant bin. 
\begin{figure*}[t]
% \vspace{-15pt}
\begin{center}
\includegraphics[width=\linewidth]{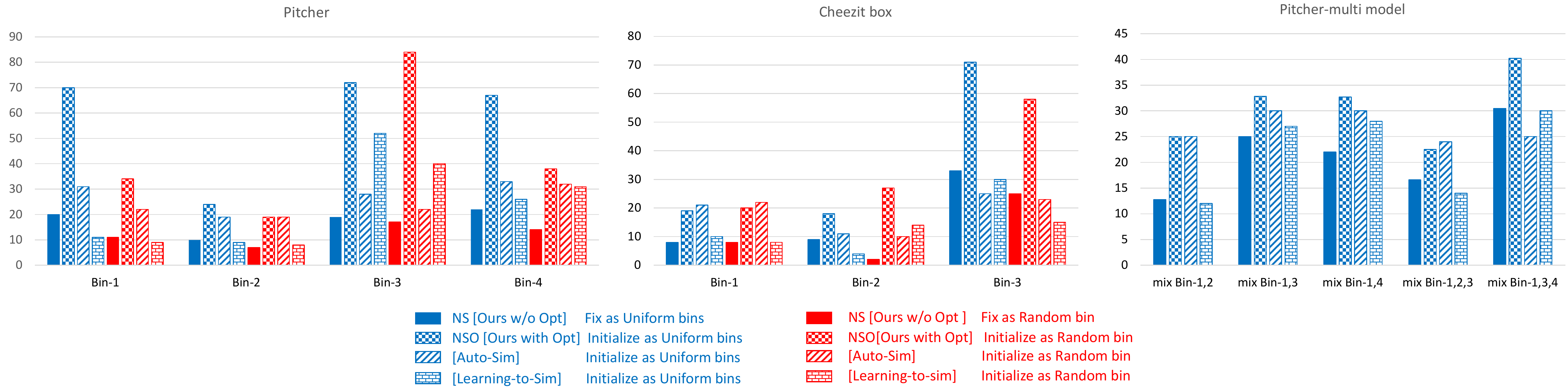}
\end{center}
% \vspace{-4mm}
   \caption{Performance of Neural-Sim on the YCB-in-the-wild dataset. We observe that the Neural-Sim optimization (NSO) can consistently achieve $20\%$ to $60\%$ improvement in accuracy over our method without optimization (NS) case and large improvements over LTS (up to 58\%) and Auto-Sim (up to 60\%). Here each bin on $x-$axis represents bin from which test data is generated. We observe large improvement in both single-modal and multi-modal test data.
   %For those objects that are sensitive with the pose (e.g., pitcher, cheezit box), compared with baseline methods that use the images to train NeRF train the object detection model, Neural-Sim improve the AP by over 80\%; Compared with our method without optimization, the optimization can lead over 50\% AP improvement. 
   }
\label{fig:wild-result}
% \vspace{-10pt}
\end{figure*}

Quantitative results are provided in the Fig.~\ref{fig:wild-result}.
%
%First, the results shows the performance of our method without optimization (fix $\psi$) where we use NeRF to generate data from uniform distribution and randomly selected bins.
%
%Next, we show how our Neural-Sim can find the optimal pose rendering parameters for both above scenarios. In the first case, we initialize the Neural-Sim optimization process with the uniform distribution and in the second case we use randomly selected bin as starting point. Our Neural-Sim then find the optimal bin distribution that are used to synthesize good training images.
%
%The performance of all experiments are shown in Fig.~\ref{fig:8}. 
%
First we highlight the performance achieved by our NS approach
%NeRF 
to generate data according two different initial pose distributions. 
We observe that NS generated data helps achieve
%to solve downstream object detection task as we observe 
up to $30\%$ in object detection accuracy on different objects starting from two different initial distributions.
%
%Taking uniform distribution $\psi_u$ as training pose distribution and comparing our methods with (blue) and without (red) optimization, 
%
Moreover, our NSO approach
%Neural-Sim approach with optimization 
achieves remarkable improvement in every experimental setup. For example, on {\em pitcher}, starting from uniform and random distributions, our optimization improve performance by almost $60\%$. Compared with other optimization methods LTS and Auto-Sim, we observe large improvement upto $58\%$ improvement over LTS and $60\%$ improvement over Auto-Sim on the pitcher object.
We observe a similar behavior on the {\em cheeze box} and also on multi-modal experiment setting.
%\neel{need to add an explanation for what is the multi-modal setting}.
%
This highlights three points. First, NeRF can be used to generate good data to solve object detection task in the wild; far more importantly, our Neural-Sim with bi-level optimization (NSO) approach can automatically find the optimal data that can help achieve remarkable improvements in accuracy on images captured in the wild. Third, the gradients from NSO are more effective and can generate better data than REINFORCE based LTS and Auto-Sim.

\subsection{YCB Video dataset}
\label{sec:5.4}

To show the performance of the proposed NS and NSO approaches 
%Neural-Sim approach 
on a general real world dataset, we also conduct experiments on the YCB-Video dataset \cite{xiang2017posecnn,hodan2018bop}. Each image in this dataset consists of multiple YCB objects (usually 3 to 6 different objects) in a real world scene. The YCB-Video training dataset consists of 80 videos from different setups in the real world. 
Since there are many duplicate frames in each video, we select every 50th frame to form the training set, which results in just over 2200 
% 8000 \AD{2227}
training images ({\em YCBV$_{train}$}).
YCB-Video testset contains 900 images. YCB-Video train and test sets have all 21 YCB objects.
%
%and 900 test images.
%
%Due to the high duplicate in each video, we down sample the dataset with sampling rate 50 and form a training set with 8000 images  $YCBV_{train}$. 
%
% \VV{What is the test set size here?} 
%
In order to show the benefit of synthetic data, we create two different training scenarios
(1) \noindent{\bf Few-shot setting}, where we randomly select 10 and 25 images from ({\em YCBV$_{train}$}) to form different few shot training sets.
(2) \noindent{\bf Limited dataset setting}, where we randomly select 1\%, 5\%, 10\% images from ({\em YCBV$_{train}$}) to form limited training sets.
% (3) \noindent{\bf Full dataset setting}, where we use full training set of 8000 images \AD{2227}. 

% \AD{combine setting}
Using a similar setting as in Sec.~\ref{sec:5.3}, we demonstrate performance of the proposed NS and NSO approaches 
%Neural-Sim approach without and with optimization 
starting from uniform distributions and compare with four baselines. First baseline-1 involves training RetinaNet using few-shot or limited training images from {\em YCBV$_{train}$} data, and baseline-2 involves training RetinaNet using the images that were used to train NeRF. Baseline-3 is Learning-to-sim and baseline-4 is Auto-Sim.
Further, we also combine the real-world few-shot or limited training images from {\em YCBV$_{train}$} along with NeRF synthesized images during our Neural-Sim optimization steps for training object detection model.
%
%
%Because in YCB-Video dataset, we have some amount of real training images (either few-shot or limited percent), besides the traditional pipeline with Neural-Sim that only use NeRF synthetic dataset to train a pretrained RetinaNet with the few-shot/limited training image, we also provide a \textit{combine} optimization setting where the limited real world training set image are involved into the optimization. 
%
%
This {\em Combined} setting reduces the domain gap between synthetic and real data. 
All the models have been evaluated on YCB-Video testset.

% %%%% Table 2
\begin{table}[b]
\begin{subtable}[c]{0.5\textwidth}
\scriptsize
\begin{center}
\begin{tabular}{c|c|c|c}
\hline
    Few-shot setting  & 0-shot & 10-shot & 25-shot  \\
\hline
Only YCBV-train      & N/A & 0.45 & 0.49  \\
 train(pre)+ours (w/o opt)           & 2.3 & 3.9  & 4.6  \\
 train(pre)+ours (with opt)                    & 4.5 & 4.9  & 4.9  \\
\hline
Learning-to-sim (com)          & N/A & 12.4 & 22.5  \\
Auto-Sim (com)                & N/A & 12.9 & 22.2 \\
train(com)+ours (w/o opt)           & N/A & 12.2 & 21.0 \\
train(com)+ours (with opt)           & N/A & \textbf{13.1} & \textbf{23.0} \\

\hline
\end{tabular}
\end{center}
\subcaption{Zero and few-shot setting (YCB-Video).}
\end{subtable}
\begin{subtable}[c]{0.5\textwidth}
\scriptsize
\begin{center}
\begin{tabular}{c|c|c|c}
\hline
    Percent of {\em YCBV$_{train}$} & 0.01 & 0.05 & 0.1  \\
\hline
%\hline
 Only YCBV-train  & 5.77 & 8.88 & 12.5 \\
 Only images to train NeRF & 3.9 & 3.9 & 3.9 \\
    %  YCBV-train + Blender No-optimization  & 0.1 & 0.1 & 0.1 & 0.1 & 0.1 & 0.1 & 0.1 & 0.1 & 0.1 \\
     train(pre)+ours (w/o opt)  & 7.9  & 11.8 & 14.4  \\
     train(pre)+ours (with opt)    & 8.9  & 12.4 & 14.5  \\
     \hline
     Learning-to-sim (com)      & 36.9 & 44.1 & 48.2  \\
     Auto-Sim (com)             & 37.1 & 43.7 & 48.3  \\
     train(com)+ours (w/o opt)  & 36.7 & 43.6 & 47.9  \\
     train(com)+ours (with opt)    & \bf{37.4} & \bf{44.9} & \bf{48.9}  \\

    %  1st Ours (w optimization) 2nd combine YCBV-train + & 0.1 & 0.1 & 0.1 & 0.1 & 0.1\\
\hline
\end{tabular}
\end{center}
% \vspace{-7pt}

\subcaption{limited data setting (YCB-Video)}
\end{subtable}
\caption{YCB-Video performance. Observe large improvement of the proposed Neural-Sim approaches before and after optimization over the baselines.}
% \vspace{-20pt}
\label{table:few-shot}
% \vspace{-5pt}
\end{table}

%
%We show results of our approach in two settings: when we optimize the data generation parameters and when we do not optimize the parameters. In both the cases, we start with uniform distributions.
%
%\noindent{\bf Few-shot setting}. We observe that when we use our Neural-Sim approach to generate data 
%
%

For the normal Few-shot setting (rows 2, 3, 4 in Tab.~\ref{table:few-shot}(a)), NS
%Neural-Sim without optimization 
%
starting from the uniform distribution achieves almost $3.45$ and $4.11\%$ improvement over the baseline in $10$ and $25$ shots settings, respectively. Further, when we optimize the parameters using NSO, we observe improvements of $4.45, 4.41\%$ over the baseline and $1.0, 0.3\%$ improvements over the NS case in $10, 25$ shot settings respectively. We also observe almost $1.8\%$ improvement in the zero-shot case.

In addition, for the \textit{Combined} Few-shot setting 
%where avoiding format domain gap 
(rows 5,6,7,8 in Table.~\ref{table:few-shot}(a)), 
we observe similar large improvements in accuracy. For example, an improvement of $22.51\%$ over the baseline and $2\%$ improvements over the without optimization cases respectively have been observed in the $25$ shot settings. Compared with Learning-to-sim and Auto-Sim, NSO shows consistent improvement on both $10$ shot and $25$ shot. 
%
%
%without optimization starting from the uniform distribution achieves almost $12.35, 20.51\%$ improvement over the baseline in $10, 25$ shots setting respectively. 
%
%Further, when we optimize the parameters, we observe an improvement of $12.55, 22.51\%$ over the baseline and $2\%$ improvements over the without optimization cases respectively in $25$ shot settings respectively. 
%
%to synthesize images used to train RetinaNet staring from unifrom distribution with and without optimization. 
% All the quantitative results are shown in Table.~\ref{table:few-shot}. 

%We observe similar performance improvement in the normal limited data settings (rows 2, 3 in Table.~\ref{table:limited}). 
%
%We observe that the proposed Neural-Sim approach without optimization achieves improvement of almost $2.13, 2.92, 1.9\%$ over the baseline in $1, 5, 10\%$ data regime. Further, we observe an improvement of almost $3.13, 3.52, 2\%$ over the baseline and $1.0, 0.6, 0.1\%$ improvements over the without optimization cases respectively in $1, 10, 25$ shot settings respectively. 
We observe similar large performance improvements in the limited data settings (Table.~\ref{table:few-shot}(b)).
For example, in the \textit{Combined} limited data settings (rows 6, 7, 8, 9 in Table.~\ref{table:few-shot}(b)), we observe that the the proposed NS achieves an improvement of almost $30.93, 34.72, 35.4\%$ over the baseline in the $1, 5, 10\%$ data regime, respectively. Further, after using NSO we observe an improvement of almost $31.63, 36.02, 36.4\%$ over the baseline.
Finally, we also find $0.7, 1.3, 1.0\%$ improvements over NS approach, $0.5, 0.8, 0.7\%$ improvements over Learning-to-sim and $0.3, 1.2, 0.6\%$ improvements over Auto-Sim
%the without optimization cases 
% respectively 
in $1, 5, 10\%$ settings respectively.
Please refer to the appendix for more results and discussion including the results on ObjectNet\cite{barbu2019objectnet} dataset.

\subsection{Interpretability of Neural-Sim}
\label{sec:5.5}

% YCB-in wild dataset + meta labeling, with our method we can generate missing bins. More interpretable show interpretable results(single model, multi-model, more details)
% use distribution to show
% \vspace{-1mm}
We have observed significant improvement in accuracy even when there exists large distribution gap between training and test sets using the proposed Neural-Sim approach.
This raises a question: does the Neural-Sim optimization provide interpretable results? 
%
%Our experiments demonstrate that the proposed Neural-Sim can learn optimal rendering parameters which can be used to synthesize data, with which one can train the detector obtaining high performance on validation set. We observe such remarkable improvement in accuracy even when there is large mismatch between training and test distributions.
%This raises a question: does the optimization approach provide interpretable results? 
%This is important for developing a reliable and trust-worthy method.
%In this section, we demonstrate that Neural-Sim synthesized data are not only efficient to lead higher accuracy, but also interpretable, which is a reliable and trust-worthy method.

%Previous experiments use different test scenarios demonstrate that Neural-Sim can learn optimal rendering parameters which can be used to synthesize on demand data, with which one can train the detector obtaining high performance on validation set.
%In this section, we demonstrate that Neural-Sim synthesized data are not only efficient to lead higher accuracy, but also interpretable, which is a reliable and trust-worthy method.
%Results highlighting that the Neural-Sim learns an interpretable output are shown in Fig~\ref{fig:10}. 

\begin{figure*}
% \vspace{-10pt}
\begin{center}
\includegraphics[width=\linewidth]{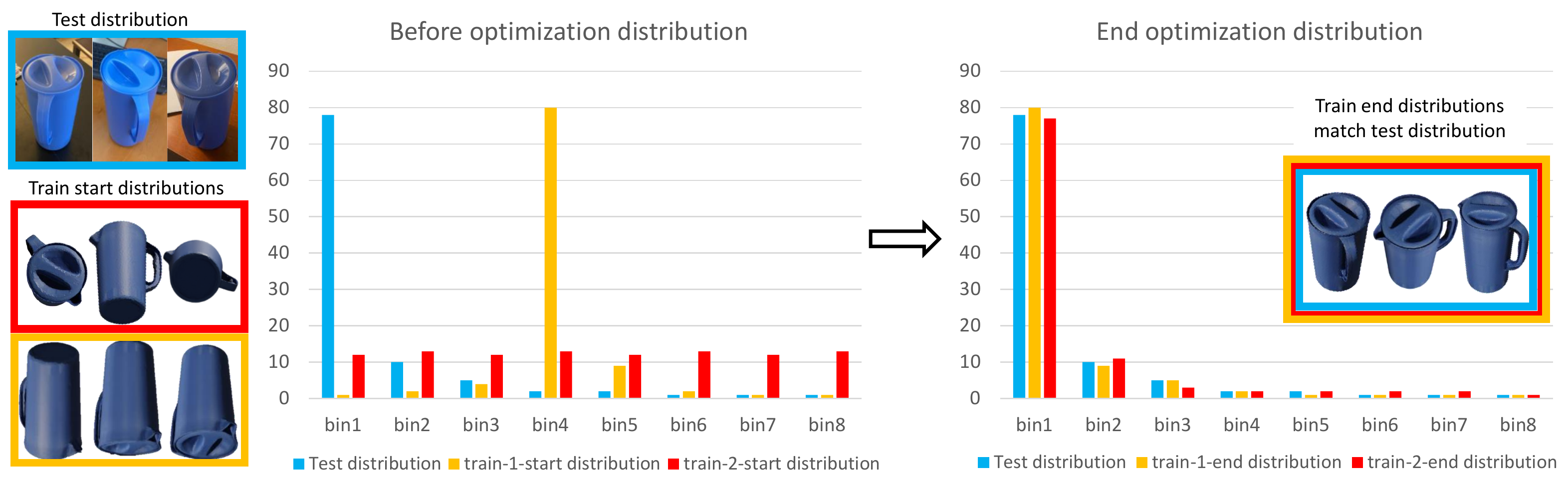}
\end{center}
% \vspace{-5mm}
   \caption{ Visualization provides evidence that proposed Neural-Sim (NSO) approach generates interpretable outputs. In the shown example, test images are sampled from distribution bin 1 as dominant bin. For Neural-Sim optimization (NSO), initial training pose distributions are uniform and bin 4 as dominant bin. Observe the bin distribution at the optimization - the final bin distribution at the end of Neural-Sim training matches with the test bin distribution.
   %[different cases. range, details, log plot, multi-model as well, traing and test] YCBV show interpretable results(single model, multi-model, more details) use distribution to show
   }
\label{fig:10}
% \vspace{-20pt}
\end{figure*}

In order to demonstrate this behavior, we conduct experiment on {\em YCB-in-the-wild} dataset illustrated in Fig~\ref{fig:10}. As shown, the test set images are sampled from the categorical distribution where bin one is dominant.
%
%%%%%%
%%%%%%   Fig-Interpretable
%%%%%%
As described in Sec.~\ref{sec:5.3}, we consider two starting pose bin distributions for our Neural-Sim approach: a uniform distribution and a randomly selected bin as a dominant bin (e.g., most images come from bin four).
%
%We optimize the pose distribution to synthesize training images $D_{train}^1$ and $D_{train}^2$ respectively for downstream object detection task.
%
%
After optimization, we visualize the learned object pose distribution (Fig~\ref{fig:10} (b)). We find that no matter what the starting distributions the Neural-Sim approach used, the learned optimal $\psi^{*}$ is always aligned with the test distribution. 
This explains the reason why Neural-Sim can improve the downstream object detection performance: it is because Neural-Sim can automatically generate data
%, which in our case, 
that will closely matching distribution as the test set. 
We can find similar interpretable results in camera zoom experiments.
More such visualization highlighting interpretable outputs are provided in the supplementary material.

    \label{sec:conclusion}
    % \vspace{-10pt}
\section{Discussion and Future Work}
% \vspace{-10pt}
It has been said that “Data is food for AI”\cite{datacentricai}. While computer vision has made wondrous progress in neural network models in the last decade, the data side has seen much less advancement. There has been an explosion in the number and scale of datasets, but the \textbf{process} has evolved little, still requiring a painstaking amount of labor.  

%In particular, the concept of ``Data-centric AI" call for the next steps in AI to be a need for attention to careful crafting of training data\cite{datacentricai}.

Synthetic data is one of the most promising directions for transforming the data component of AI.  While it has been used to show some impressive results, its wide-spread use has been limited, as creating good synthetic data still requires a large investment and specialized expertise. 

We believe we have taken a big step towards making synthetic data easier to use for a broader population.  By optimizing for how to synthesize data for training a neural network, we have shown big benefits over current synthetic data approaches.  We have shown through extensive experiment that the data found by our system is better for training models. We have removed the need for any 3D modeling and for an expert to hand-tune the rendering parameters. This brings the promise of synthetic data closer for those that don’t have the resources to use the current approaches.

We have handled camera pose, zoom and illumination; and our approach can be extended to other parameters (such as materials, etc.), by incorporating new advances in neural rendering. For future work,
we hope to improve the ease of use of our approach, such as performing our optimization using lower quality, faster rendering using a smaller network for the neural rendering component, and then using the learned parameters to generate high quality data to train the final model.  We hope that our work in this space will inspire future research.

%\neel{I'd like to take a pass at this in a bit.  I like the sentiment, but I think it could be stronger.  The Andrew Ng quotes are helpful, but it sort of puts him as the expert and as opposed to us :).  Need to think a little how to use this but reframe a bit.}

% \noindent{ \bf Human effort comparison}
% \begin{enumerate}
% \item Different approximation (in controllable experiments, algorithm properties)
% \end{enumerate}

% \begin{enumerate}
% \item Compare with Bayesian optimization, random search.
% \end{enumerate}

\textbf{Acknowledgments} We want to thank Yen-Chen Lin for his help on using the nerf-pytorch code. This work was supported in part by C-BRIC (one of six centers in JUMP, a
Semiconductor Research Corporation (SRC) program sponsored by DARPA),
DARPA (HR00112190134) and the Army Research Office (W911NF2020053). The
authors affirm that the views expressed herein are solely their own, and
do not represent the views of the United States government or any agency
thereof.

\clearpage
\bibliographystyle{splncs04}
\bibliography{neural-sim-arXiv}

\clearpage
\appendix
\section*{Appendix}

We provide additional information about the implementation details (Sec. \ref{sec:implementation}), different datasets used (Sec. \ref{sec_supp:dataset}), and experimental results (Sec. \ref{sec_supp;experiments}).

\section{Implementation Details}
\label{sec:implementation}

\subsection{Memory Efficiency}

\paragraph{Tool 2: Twice-forward-once-backward} \label{twice}

As discussed in the main paper Sec.~3.1 (Tool2), the full gradient update of our bi-level optimization problem involves using the approximation of $\nabla_{NeRF}$ in Eq. 5 and back in Eq. 2. There are three terms in this computation with the following dimensions: 

\noindent 
\begin{inparaenum}[(1)]
\item $\frac{\partial(\frac{\partial l(x_j, \hat{\theta}(\psi_{t}))}{\partial\theta})}{\partial x_j} \in \mathbb{R}^{m\times d}$,
\item $\frac{\partial x_j}{\partial\psi} \in \mathbb{R}^{d\times k}$, and 
\item $\nabla_{TV}=\mathcal{H}(\hat{\theta}(\psi_{t}), \psi)^{-1}\frac{d\mathcal{L}_{val}(\hat{\theta}(\psi_t))}{d\theta}\in \mathbb{R}^{m\times 1}$,
\end{inparaenum}
where $m=|\theta|$ is the $\#$ of parameters in object detection model, $d$ is the $\#$ of pixels in $x$, and $k$ is $\#$ of pose bins. 

Specifically, if we follow the sequence of $(3)$-$(1)$-$(2)$, first, the output of $(3)$ is a $1\times m$ vector and can be used as the weight to compute $(1)$ with Pytorch weighted autograd. In this case, we do not need to explicitly store the huge matrix ($\mathbb{R}^{m\times d}$) of $(1)$ and the corresponding large computation graphs of each element in it. Similarly, the result of previous step ($(3)$-$(1)$) is a $1\times d$ dimension vector and we can use it as the weight to compute $(2)$ with Pytorch weighted autograd. Finally, we obtain the gradient as a $1\times k$ dimension vector.

\paragraph{Tool 3: Patch-wise gradient computation}
\label{patch-wise}

As discussed in the main paper Sec. 3.1 (Tool3), patch-wise gradient computation helps to save the memory cost of computing $(1)$-$(2)$ sequence of the gradient. Table.~\ref{table:patch} shows the details about the memory cost when we use different patch sizes. Specifically, if we keep NeRF chunk as 512 fix, when we increase the patch size by multiplying 2 each time, the memory cost of gradient computation will also approximately doubled.
With Patch-wise gradient computation, image size would not be the bottleneck of gradient computation.

%%%% Table1
\begin{table}
% \small
\scriptsize
\begin{center}
\begin{tabular}{c|c|c}
\hline
 Patch Size  & Gradient Computation Memory Cost & Total Memory Cost \\
\hline
 512 (32x16) & 1910 MB & 6450 MB  \\
 256 (16x16)  & 974MB & 5514MB   \\
 128 (16x8) & 464MB & 5004MB  \\
 64 (8x8) & 216MB & 4756MB  \\

\hline

\end{tabular}
\end{center}
\caption{Patch-wise optimization memory cost comparison with different patch-size, for the total memory cost we fix the NeRF chunk as 512.}
\label{table:patch}
\end{table}

% \subsection{NeRF training}

% How to train NeRF and the property of NeRF (effect of near and far, zoom factor, image quality with # rays)

% \subsection{Generation of ground truth data}

% To get the ground truth, we modify NeRF output and automatically obtain the groundtruth

\subsection{Comparison with the graphics pipeline}

% More results of Figure.5 experiments.

\begin{figure}
\begin{center}
\includegraphics[width=\linewidth]{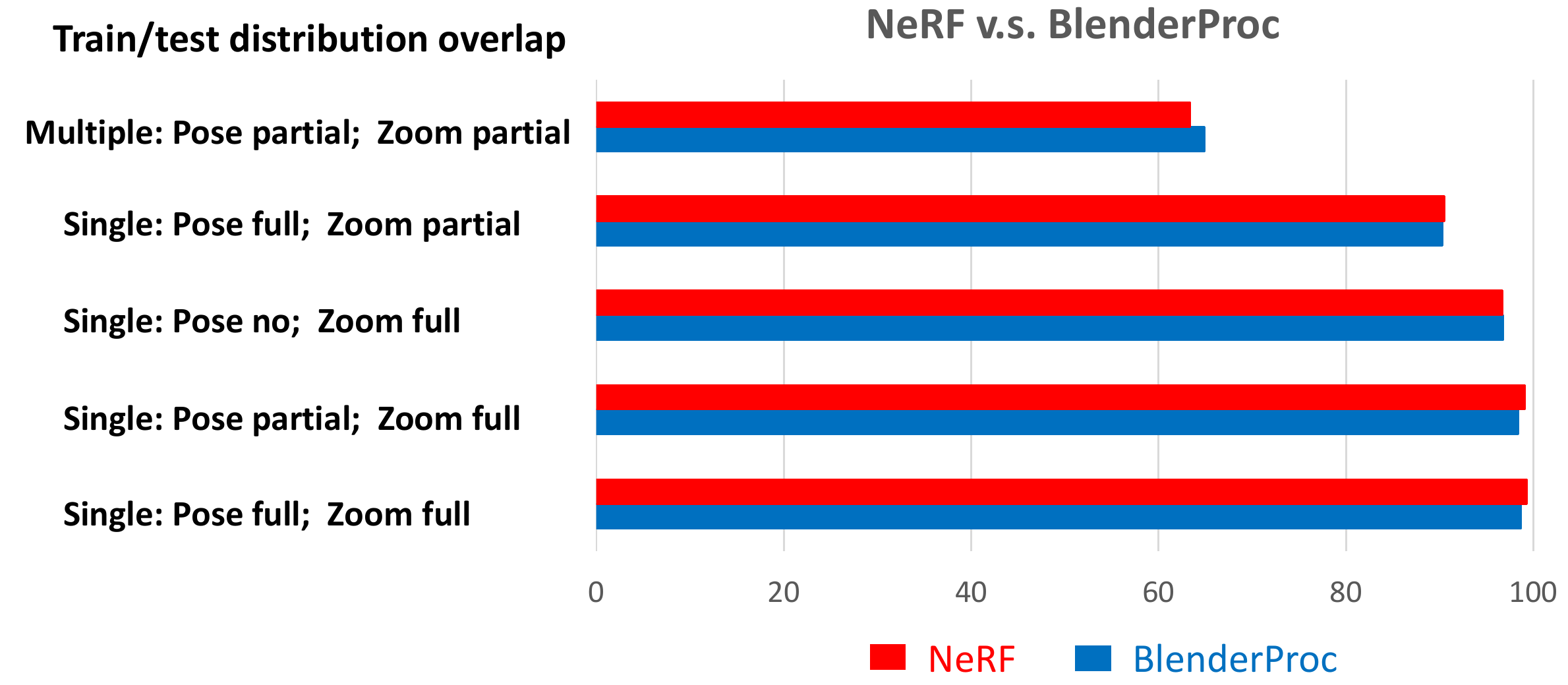}
\end{center}
  \caption{Object detection performance using NeRF and Blender synthesized images. The X-axis is mAP. Single means each test image contains one object, multiple means each test image contains multiple objects.}
\label{fig:nerf_blemder}
\end{figure}

We provide additional quantitative results to demonstrate that NeRF can replace traditional graphics pipelines like BlenderProc \cite{denninger2019blenderproc} when generating data for downstream computer vision tasks such as object detection. Results are provided in the Fig. \ref{fig:nerf_blemder}. 

%
%First, it is important to show that NeRF is a suitable replacement for a traditional renderer like BlenderProc \cite{denninger2019blenderproc} when generating data for object detection. 
%
%
%In order to create YCB-synthetics dataset, 
To test this, we consider objects from YCB-video datasets. 
%We selected these objects as they show large appearance variance under rendering.  
%
We render images from NeRF and BlenderProc \cite{denninger2019blenderproc} using the same camera pose and zoom parameters.
We use these images to conduct multi-class object detection tasks under the same training and test setting. 
As shown NeRF generated data can achieve the same accuracy as that of BlenderProc on downstream object detection tasks.

%%%%%%
%%%%%%   Fig-7
%%%%%%

\begin{figure*}
\begin{center}
\includegraphics[width=\linewidth]{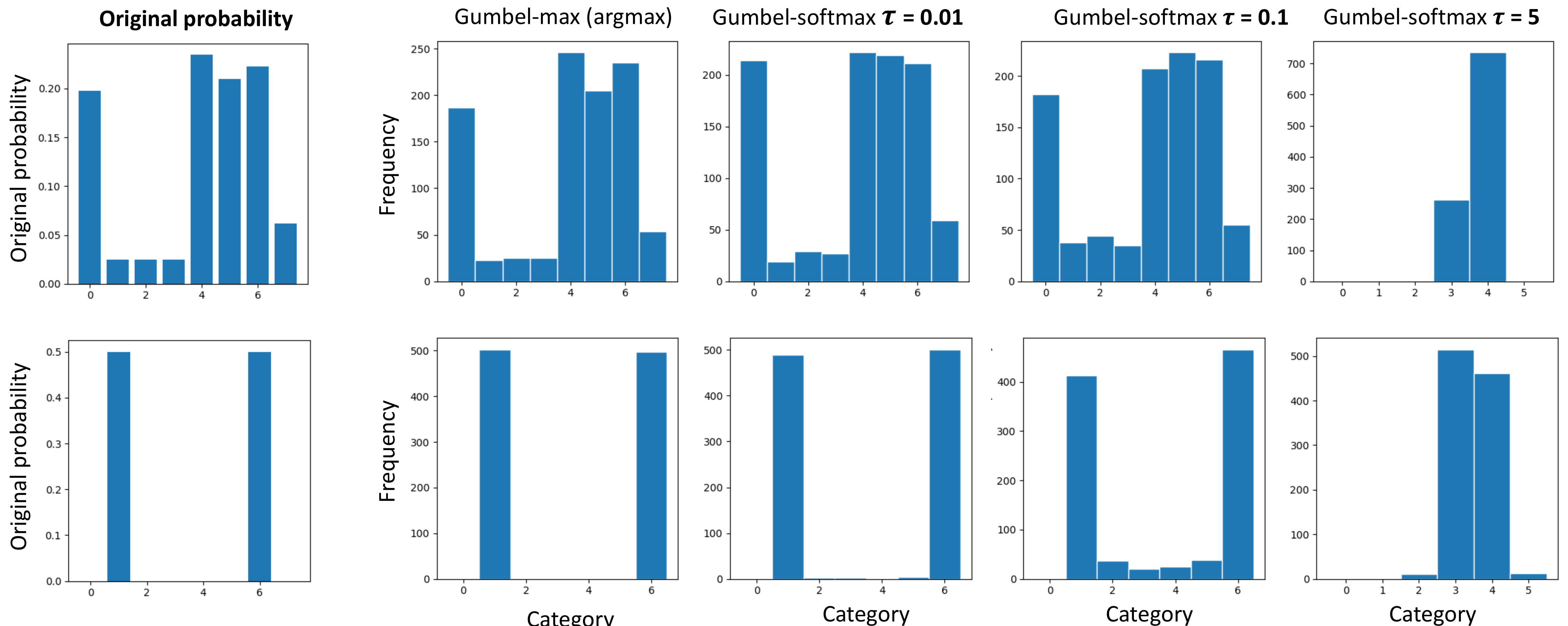}
\end{center}
  \caption{Effect of Gubmel softmax temperature parameter used to approximate the initial distribution. }
\label{fig:GS}
\end{figure*}

\subsection{Influence of optimization parameters}

% Influence of Optimization parameters: Softmax temperature, Gumble softmax(plots results) temperature. Stability trick (lr, strategy (sgd, mometum, adam), ).

We briefly describe the effects of different parameters used in our bilevel optimization updates. In particular, we show the effect of Gubmel softmax temperature parameter used in the main paper. Fig.~\ref{fig:GS} shows the Gaumble softmax performance under different temperatures. If the temperature parameter is very large, initial categorical distribution takes form of uniform distribution after Gumbel updates and
% (e.g., when temperature is 5, the output vector $y$ is similar as uniform, after times the bin degree to select bin, it will average all bin degrees which is the center degrees: bin3 and bin4).
at a lower temperature, the distribution becomes peaky. In our experiments, we have used a parameter value of $0.1$. 
%maintains the initial distribution behavior.

We used stochastic gradient descent with momentum for $\psi$ parameter updates with learning rates of 1$e$-5 and momentum value of 0.9 value. Further, on $YCB-video$ dataset, we use warm start to conduct experiment. 

\subsection{Optimization Runtime}

We now provide running time details of our end-to-end pipeline. Each iteration involves data generation through NeRF, detection model training, backpropagation through detection model including hessian-vector product evaluation, and backpropagation through data generation process. For $YCB-synthetics$ experiments described in Sec.4.2 in the main paper, it takes roughly ten minutes to complete one end-to-end computation. Further, time depends on the image resolution generated by NeRF, and detection model training. 
%
%It takes roughly twenty minutes to 
%

Finally, it should be noted that this pipeline does not involve any human effort. In comparison, the traditional graphics pipeline will involve human expert involvement for creating good 3D models of objects.

%image resolution, 20 min for each iteration, including:
%no human involvement

\subsection{Rendering from SFM}

\begin{figure}
\begin{center}
\includegraphics[width=\linewidth]{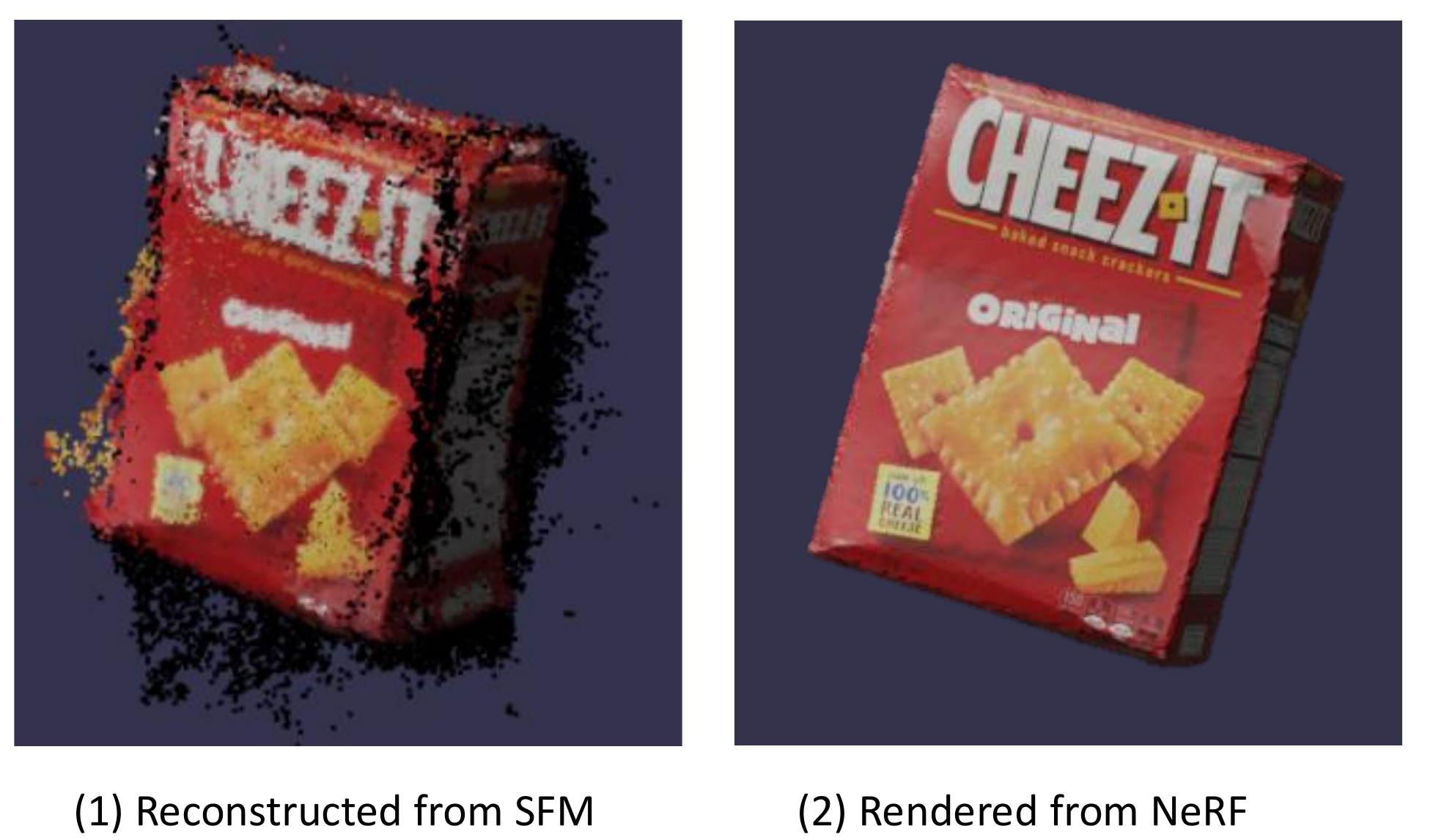}
\end{center}
  \caption{Quality of images rendered using 3D model generated using a standard structure-from-motion pipeline. In comparison, NeRF can generate high-quality images without needing high-quality 3D models.}
\label{fig:supp-Fig-SFM}
\end{figure}

In order to generate images from a traditional graphics pipeline, one needs to have accurate 3D object models including accurate geometry, texture, materials of objects. Capturing these accurate properties of the objects requires human expert involvement. 
However, if we use a standard computer vision pipeline like structure-from-motion \cite{Schonberger_2016_CVPR} pipeline to generate 3D models, the quality of images generated by these models are not as high as that of NeRF. Please refer to the Fig. \ref{fig:supp-Fig-SFM}. 
Thus involving human experts to improve 3D model quality for traditional graphics limits their scalability, and is also expensive. In contrast, NeRF only requires images along with camera pose information, providing benefits over traditional graphics pipelines. 

\begin{figure*}
\begin{center}
\includegraphics[width=\linewidth]{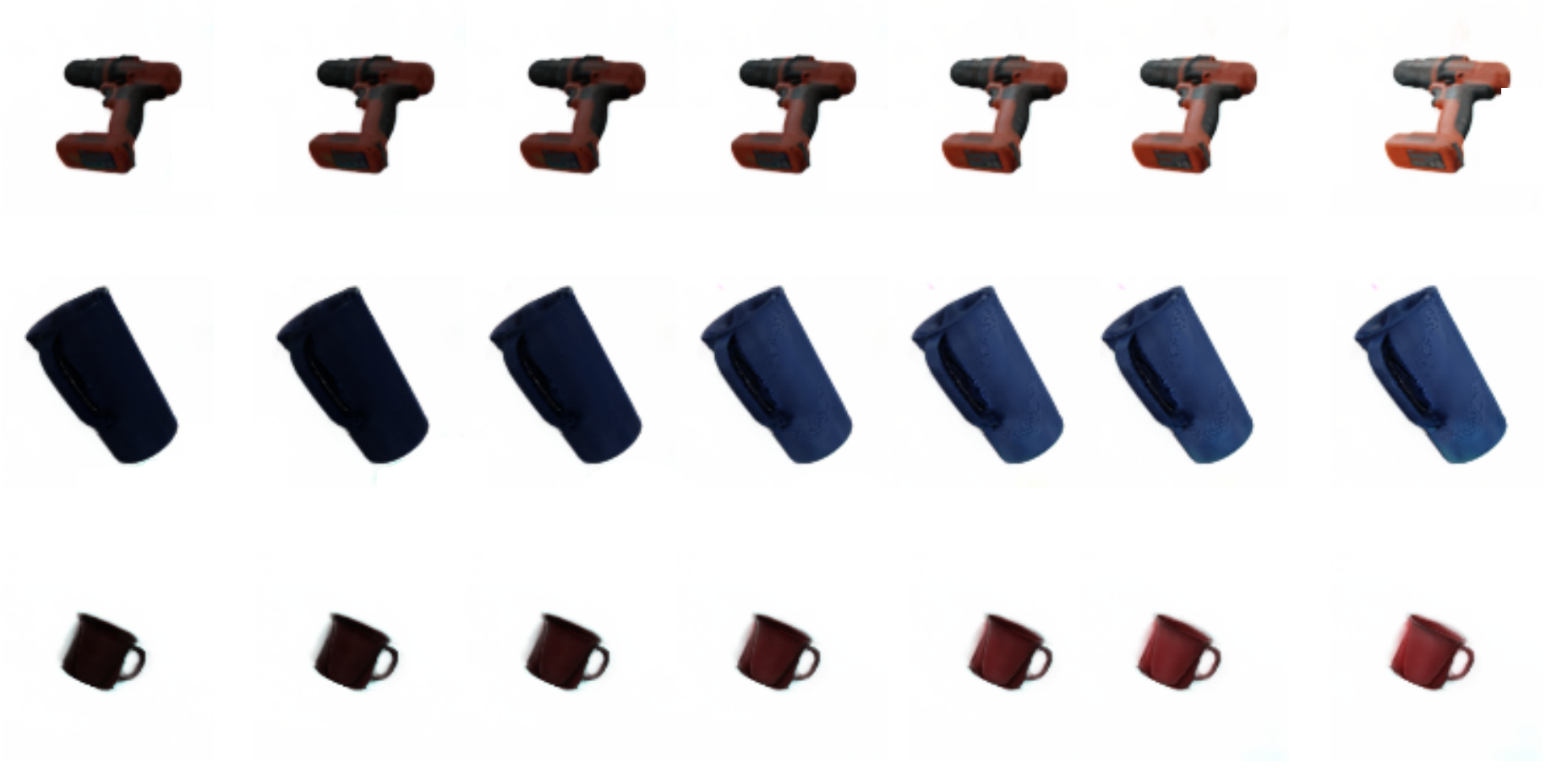}
\end{center}
  \caption{NeRF-in-the-wild illumination rendering control on YCB objects. Interpolations between the appearance embedding of two training images (left, right), which results in rendering (middle) where illuminations are interpolated but geometry is fixed. }
\label{fig:nerf-w}
\end{figure*}

\subsection{NeRF-in-the-wild}

Fig.~\ref{fig:nerf-w} shows the results of NeRF-in-the-wild (nerf-w) on controllable illumination change which allows smooth interpolations between color and lighting. The experiments have been conducted on the YCB-objects. For each object, we conduct interpolations between the appearance embedding of two training images (left, right), which results in rendering (middle) where illumination are interpolated but geometry is fixed.

\section{Dataset Information}
\label{sec_supp:dataset}
\subsection{YCB object visualization}

\begin{figure}[h]
\begin{center}
\includegraphics[width=\linewidth]{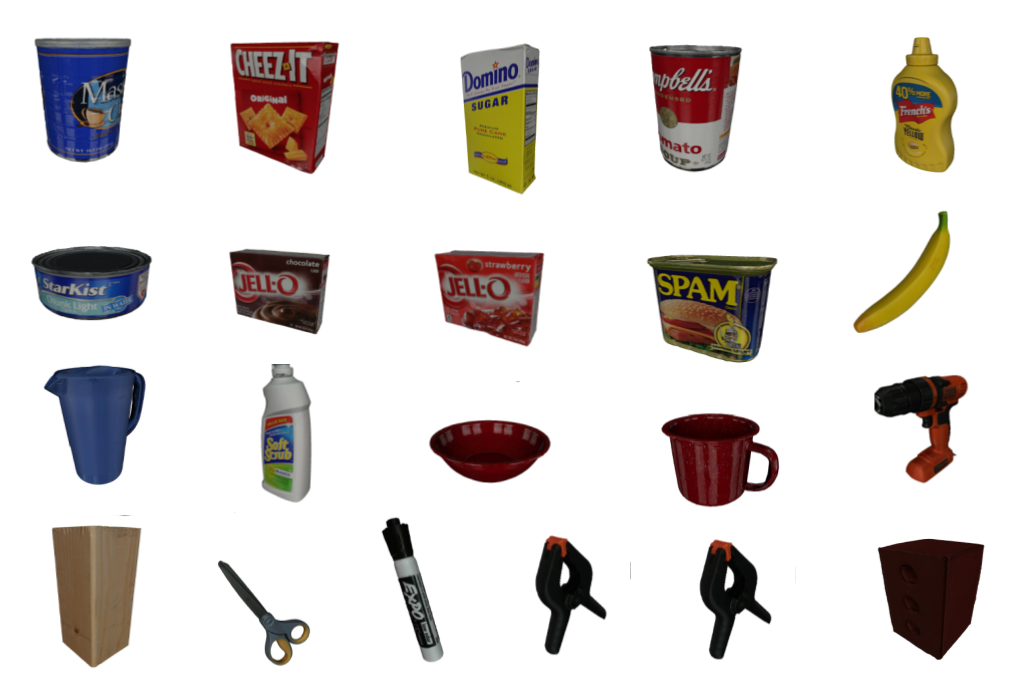}
\end{center}
  \caption{Object from YCB-video dataset \cite{xiang2017posecnn}. {\em master chef can, cracker box, sugar box, tomato soup can, mustard bottle, tuna fish can, pudding box, gelatin box, potted meat can, banana, pitcher base, bleach cleanser, bowl, mug, power drill, wood block, scissor, large marker, large clamp, extra large clamp, foam brick}}
\label{fig_supp:ycb_objects}
\end{figure}

Experiments have been conducted on $21$ objects from YCB-video datasets. These objects are: {\em master chef can, cracker box, sugar box, tomato soup can, mustard bottle, tuna fish can, pudding box, gelatin box, potted meat can, banana, pitcher base, bleach cleanser, bowl, mug, power drill, wood block, scissor, large marker, large clamp, extra large clamp, foam brick}. These objects are visualized in Fig. \ref{fig_supp:ycb_objects}.

\subsection{YCB synthetic dataset details}

% YCB synthetic dataset details (use BlenderProc synthesize image to train NeRF)

% Include instructions for data capture.

In order to train NeRF, we first use 3D YCB object models from the BOP-benchmark page \cite{hodan2018bop}. We use BlenderProc \cite{denninger2019blenderproc} to generate $100$ images per object. These images are captured from poses that are sampled from a uniform distribution. These images along with their corresponding pose values are used to train NeRF. NeRF training takes almost $20$ hours for each object.

\subsection{YCB-in-the-wild dataset}
% YCB-in-the-wild dataset details and visualization

\begin{figure*}[h]
\begin{center}
\includegraphics[width=\linewidth]{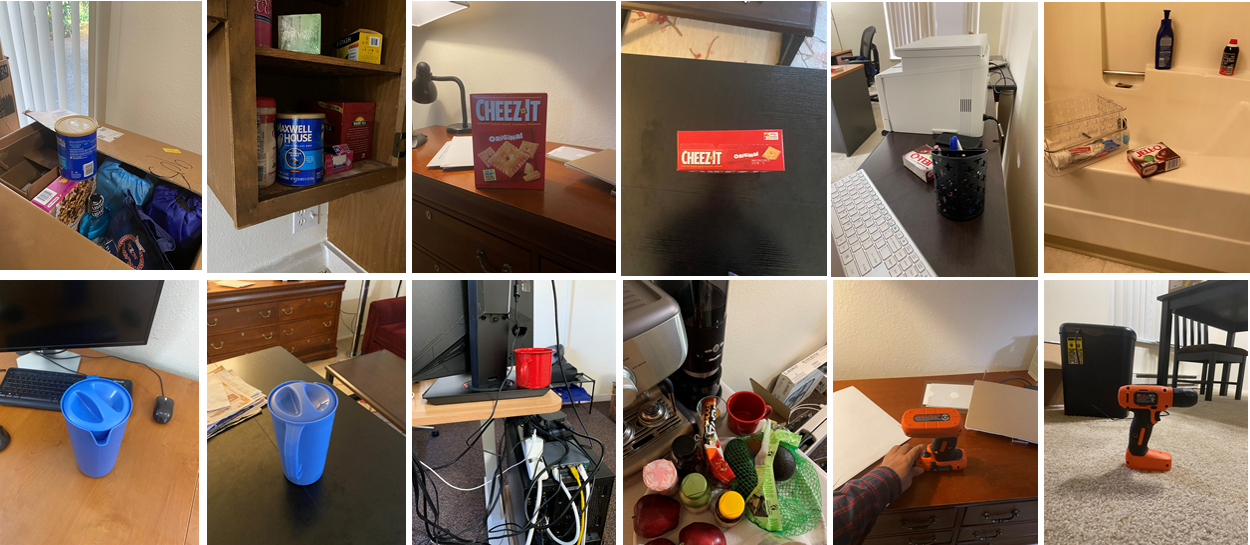}
\end{center}
  \caption{Images from our YCB-in-the-wild dataset. It consists of six YCB objects {\em masterchef, cheezeit, gelatin, pitcher, mug, driller} captured in common indoor environments: living room, kitchen, bedroom, bathroom.}
\label{fig:supp-vis-wild}
\end{figure*}

As described in the main paper, in order to evaluate the performance of the proposed NS and NSO approaches on a real-world dataset, we have created a
real-world YCB-in-the-wild dataset. The dataset has 6 YCB
objects in it, which are the same as in the YCB-synthetic dataset:
{\em masterchef, cheezit, gelatin, pitcher, mug, driller}. All
images are captured using a smartphone camera in common
indoor environments: living room, kitchen, bedroom
and bathroom, under natural pose and illumination. Images from the dataset capturing different environment properties are shown in Fig. \ref{fig:supp-vis-wild}.

%%%%%%
%%%%%%   Fig-supp-autosim-zoom
%%%%%%
% \begin{figure}[h!]
% \begin{center}
% \includegraphics[width=\linewidth]{Fig_supp/supp-Fig-posebias-opt-2.pdf}
% \end{center}
%   \caption{Comparison of the proposed NSO with the baseline Auto-Sim approach on zoom experiments conducted on on YCB-Synthetic dataset. When there are zoom distribution gaps between train and test sets, with the increase in the gap, object detection performances drops (solid line). With the help of Neural-
% Sim (NSO), the performance drops are filled. Observe large improvement of NSO over Auto-Sim.}
% \label{fig:supp-autosim-zoom}
% \end{figure}

\subsection{YCB-video dataset}

\begin{figure*}[h]
\begin{center}
\includegraphics[width=\linewidth]{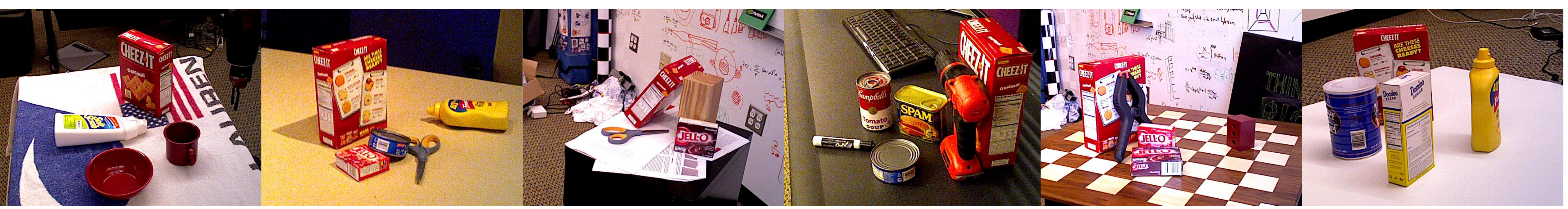}
\end{center}
  \caption{ Example images from the YCB-Video test set.}
\label{fig:supp-vis-video}
\end{figure*}

Images from the YCB-video dataset captured in different scenes are shown in Fig. \ref{fig:supp-vis-video}.

\section{Additional Experiments}
\label{sec_supp;experiments}

We provide additional experiments below.

\subsection{Interpretable experiments visualization}

\begin{figure*}
\begin{center}
\includegraphics[width=\linewidth]{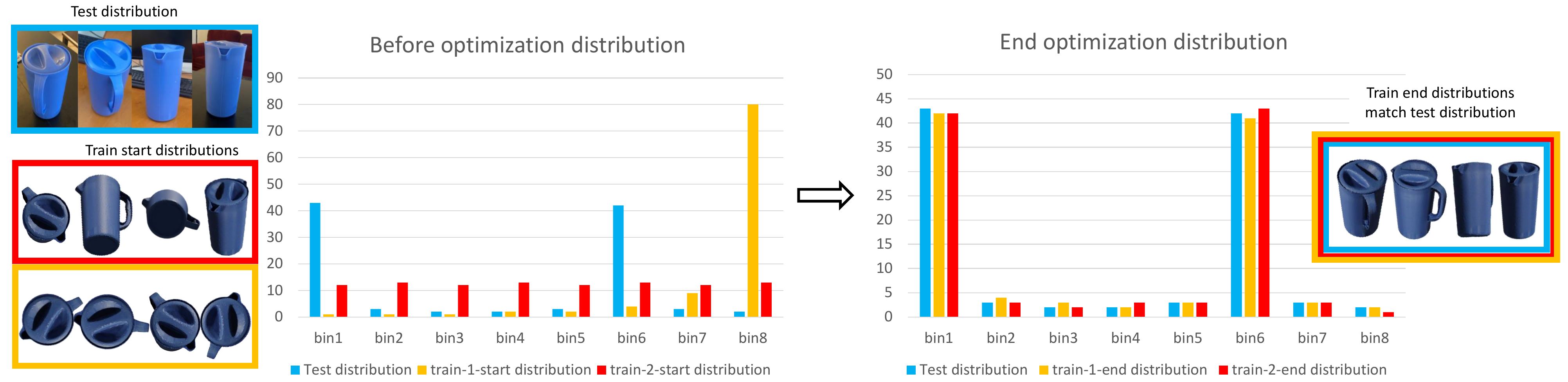}
\end{center}
  \caption{Interpretability results on multi-modal test distributions. Visualization provides evidence that the proposed Neural-Sim (NSO) approach generates interpretable outputs. In the shown example,
test images are sampled from multi-modal distribution bin 1 and bin 6 as dominant bins. For Neural-Sim optimization (NSO), initial training pose distributions
are uniform and bin 8 as dominant bin. Observe the bin distribution at the optimization - the final bin distribution at the end of Neural-Sim (NSO)
training matches with the test bin distribution.}
\label{fig:supp-Fig-inter-multi-1}
\end{figure*}

\begin{figure*}
\begin{center}
\includegraphics[width=\linewidth]{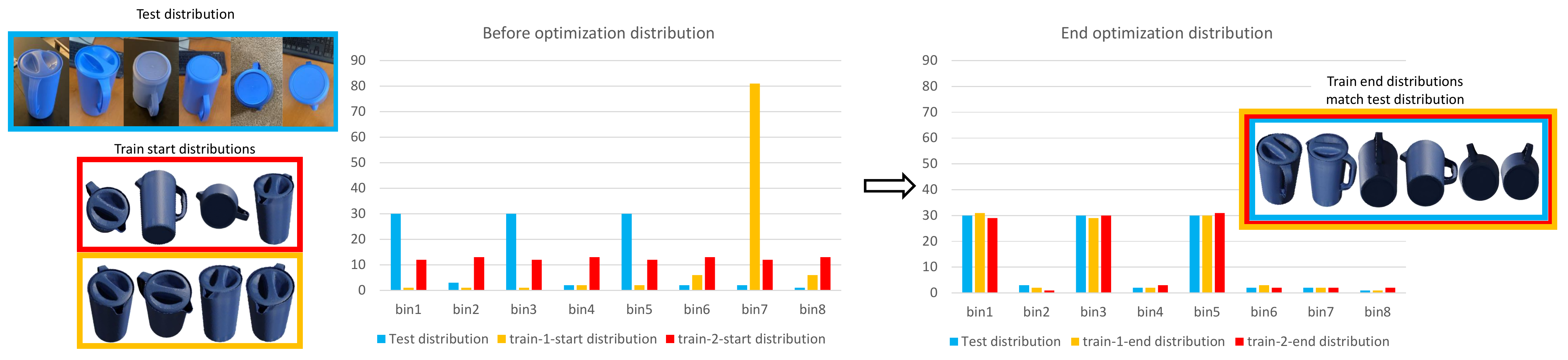}
\end{center}
  \caption{Interpretability results on multi-modal test distributions. Visualization provides evidence that the proposed Neural-Sim (NSO) approach generates interpretable outputs. In the shown example,
test images are sampled from multi-modal distribution bin 1, bin 3 and bin 5 as dominant bins. For Neural-Sim optimization (NSO), initial training pose distributions
are uniform and bin 7 as dominant bin. Observe the bin distribution at the optimization - the final bin distribution at the end of Neural-Sim (NSO)
training matches with the test bin distribution.}
\label{fig:supp-Fig-inter-multi-2}
\end{figure*}

\begin{figure*}
\begin{center}
\includegraphics[width=\linewidth]{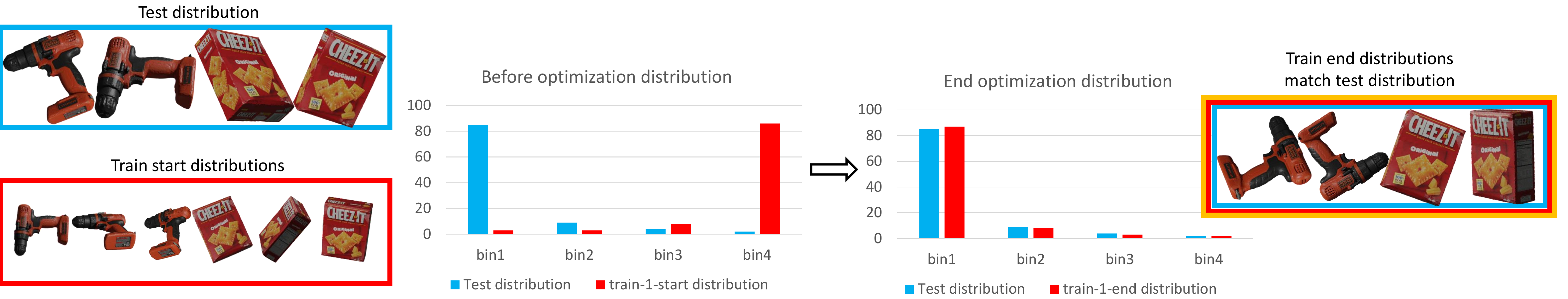}
\end{center}
  \caption{Interpretability results on zoom test distributions. Visualization provides evidence that the proposed Neural-Sim (NSO) approach generates interpretable outputs on driller and cheeze  box. Observe the bin distribution at the end of the optimization - the final bin distribution at the end of Neural-Sim (NSO) training matches with the test bin distribution.}
\label{fig:supp-Fig-inter-zoom-1}
\end{figure*}

In order to support the claim that the proposed Neural-Sim Optimization (NSO) approach can learn interpretable results, we provide additional visualization on the YCB-in-the-wild dataset illustrated in Fig. \ref{fig:supp-Fig-inter-multi-1}, Fig. \ref{fig:supp-Fig-inter-multi-2} and Fig. \ref{fig:supp-Fig-inter-zoom-1}. In particular, we demonstrate interpretability of our method on two scenarios. First, we conduct experiments where test images are generated from multi-modal distributions - two modal and three modal distributions. Second, we also show results on zoom experiments on cheeze box and driller objects. In both these experiment setup, we consider two starting bin distributions for training: a uniform distribution and a randomly selected bin as a dominant bin. As shown in the figure, we observe that no matter what the starting training distributions our NSO approach use, the final learnt distributions match with the test distributions. 

\begin{figure*}
% \vspace{-25pt}
\begin{center}
\includegraphics[width=\linewidth]{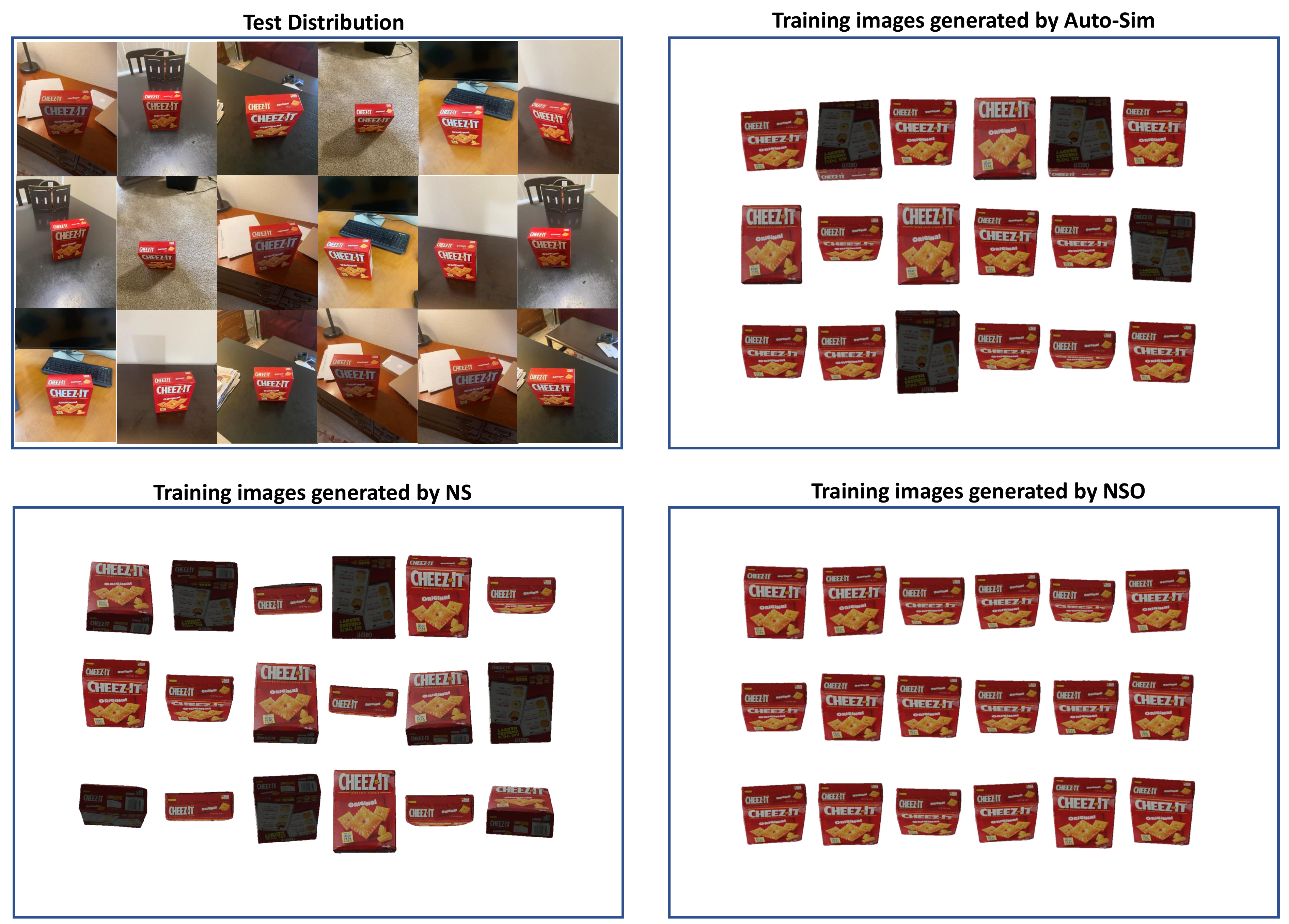}
\end{center}
  \caption{Visualization of cheeze box images generated from the learned distributions using the proposed NSO approach and the baseline Auto-Sim approach. Observe how images generated from our approach align with the test distribution. In comparison, there are many noisy samples in the Auto-Sim approach.}
\label{fig:supp-Fig-dis-1}
\end{figure*}

\begin{figure*}
% \vspace{-25pt}
\begin{center}
\includegraphics[width=\linewidth]{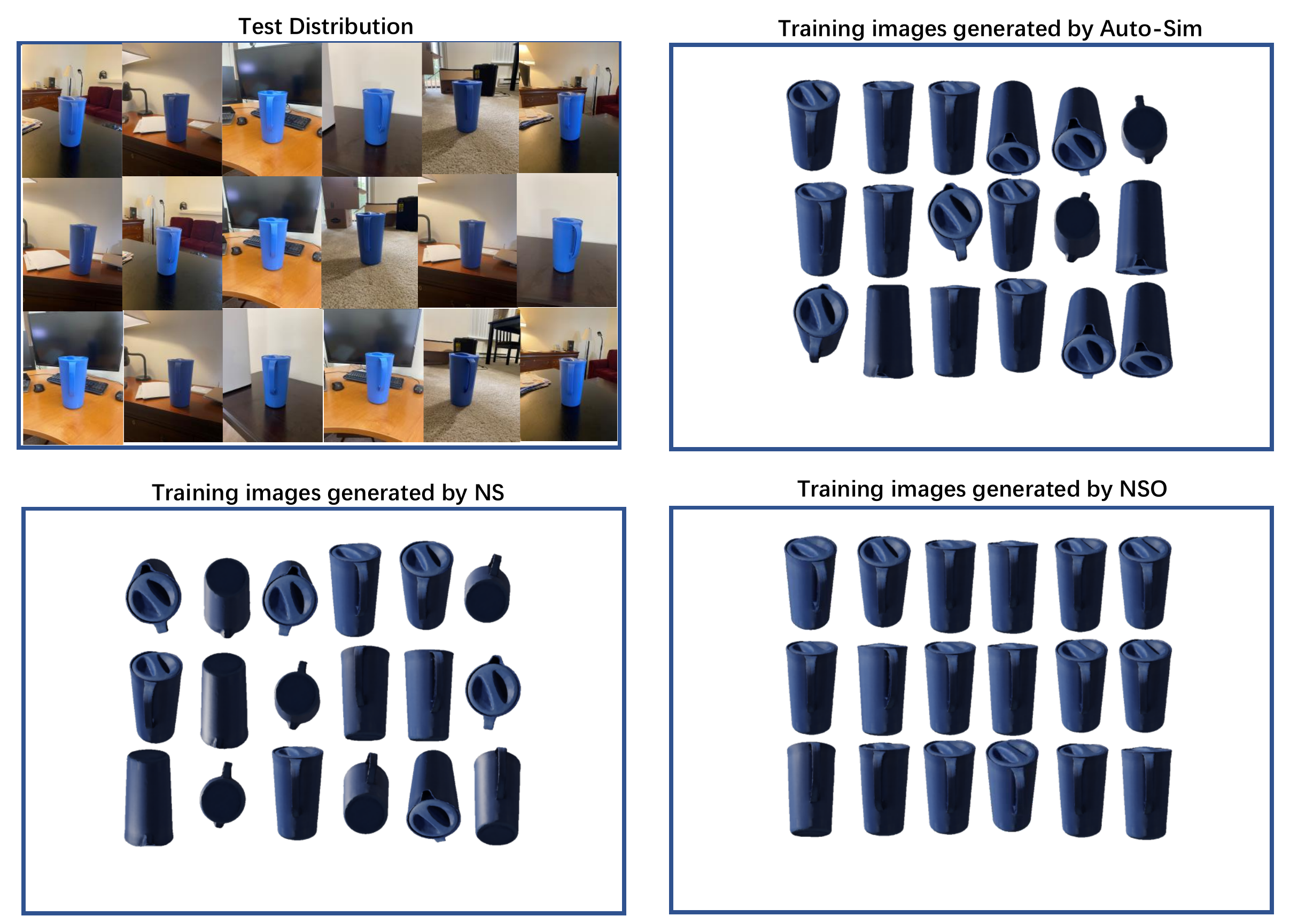}
\end{center}
  \caption{Visualization of pitcher images generated from the learned distributions using the proposed NSO approach and the baseline Auto-Sim approach. Observe how images generated from our approach align with the test distribution. In comparison, there are many noisy samples in the Auto-Sim approach.}
\label{fig:supp-Fig-dis-2}
\end{figure*}

Finally, we also provide qualitative comparison between images generated from the learned distributions using the proposed NSO approach and the baseline Auto-Sim approach. Fig. \ref{fig:supp-Fig-dis-1} provides visualization for {\em Cheeze box} and Fig. \ref{fig:supp-Fig-dis-2} for {\em pitcher} object. We have visualized eighteen images sampled from the learned distributions from the NSO and Auto-Sim approaches. We observe that the proposed NSO approach can generate images that resemble the test images in both these objects. However, Auto-Sim generates images where objects are not always aligned with test images. These visualizations provide ample evidence to support the significant improvement in performance achieved by our NSO approach over the baseline.

%%%%%%
%%%%%%   Fig-supp-Fig-inter-multi-1
%%%%%%

%%%%%%
%%%%%%   Fig-supp-Fig-inter-multi-2
%%%%%%

%%%%%%
%%%%%%   supp-Fig-inter-zoom-1
%%%%%%

%We can observe that in both s

%We also visualize and compare images generated by the propose NSO approach and baselin Auto-Sim approach at the 

%This is also highlighted from the visualization of training images generated from the final learnt distribution in Fig. \ref{}. 

%- our NSO approach automatically generates data closely resembles those of test images.

% Interpretable experiments visualization (multi-model, gaussian).

%%%%%%
%%%%%%   supp-Fig-dis-1
%%%%%%

%%%%%%
%%%%%%   supp-Fig-dis-2
%%%%%%

\subsection{Detection visualization}

\begin{figure}[h]
\begin{center}
\includegraphics[width=\linewidth]{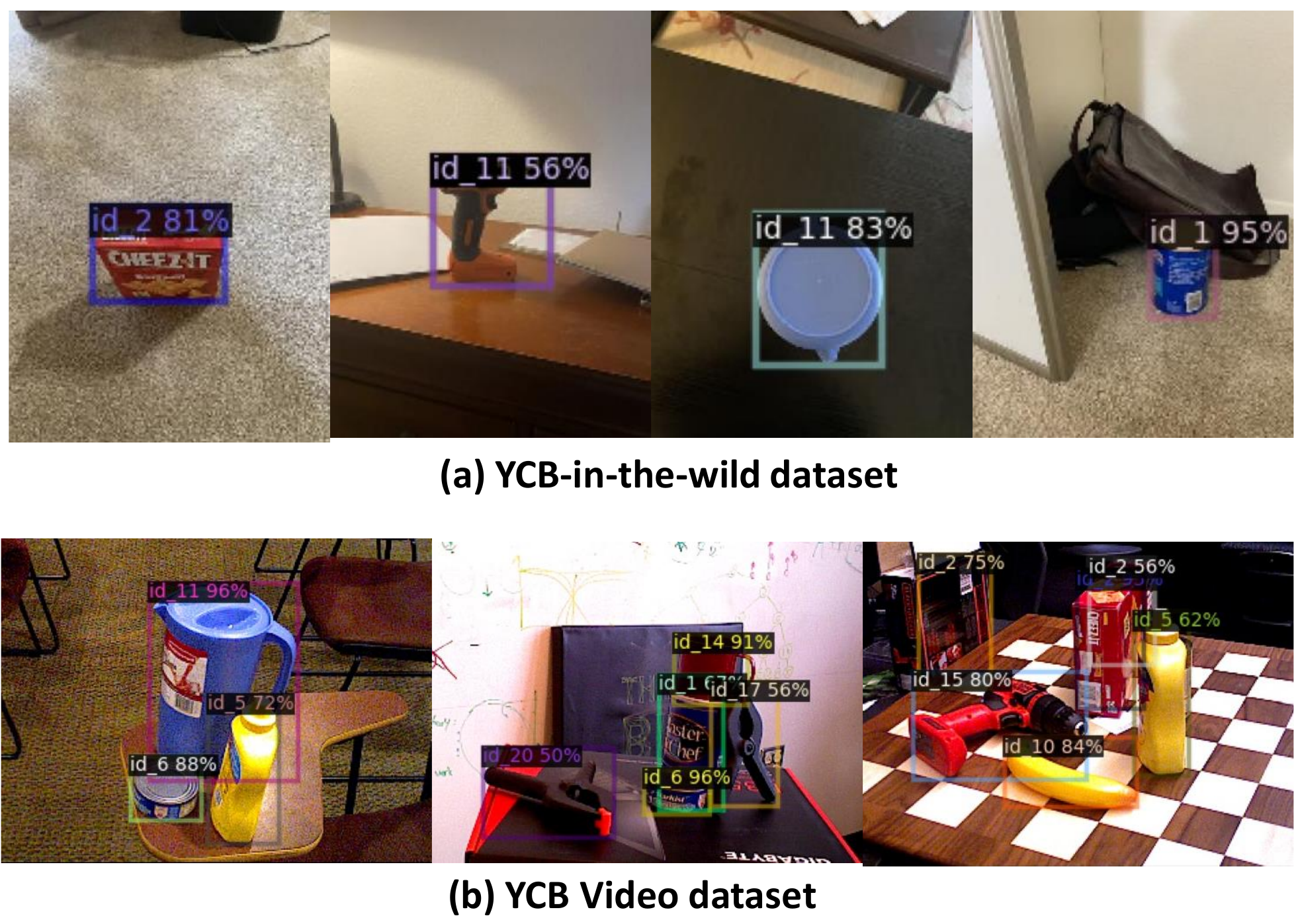}
\end{center}
% \vspace{-5mm}
   \caption{Visualization for detection results on YCB-in-the-wild and YCB-Video datasets.}
\label{fig:supp_obj_det_vis}
% \vspace{-15pt}
\end{figure}

Object detection results from our pipeline on YCB-in-the-wild and YCB-video datasets are shown in Fig. \ref{fig:supp_obj_det_vis}.  

%%%%%%
%%%%%%   Fig-visualization
%%%%%%

\subsection{Full YCB-Video dataset results}

If we use 100\% full YCB-Video training images to train RetinaNet, the mean Average Precision (mAP) reaches 58.5\% on all 21 classes.  After we use NSO with combine optimization which combines both real-world training images and NeRF synthesized images into training and optimization, the accuracy can improve from 58.3\% to 58.8\%.

\subsection{Extension to ObjectNet dataset}
We conduct experiments on ObjectNet \cite{barbu2019objectnet} dataset, which is a large real-world dataset for object recognition. The dataset consists of 313 object classes with 113 overlapping ImageNet classes. In order to synthesize training images, we use CO3D \cite{reizenstein2021common} dataset
that provides data to train NeRF. There are 17 classes that overlap with ImageNet and ObjectNet classes. After we
trained NeRFs for these classes, we find by using NeRF synthesized data to finetune an ImageNet pretrained model
provides a 4\% improvement on the 17 ObjectNet classes.
% \subsection{videos}

% \subsection{Object specific numbers}

% \subsection{YCB-in-the-wild whole dataset}

\section{Limitations and Dataset copyright}

% Limitations and Dataset copyright, societal impact.

In this work, we have focused on optimizing camera pose, zoom factor, and illumination parameters. In the real world, there are other scene parameters that affect accuracy, like materials, etc. However, our approach can be extended to include other parameters by incorporating
new advances in neural rendering. 

\paragraph{Dataset copyright.} We used publically available data. The YCB-Video dataset is released under the MIT License. Further, we will release our YCB-in-the-wild dataset under the creative commons license.

\paragraph{Societal impact} Our work focuses on using neural rendering for generating images for solving downstream computer vision tasks. This provides an opportunity to reduce reliance on human or web-captured training data that has potential privacy issues. 

\clearpage

% \clearpage
% % ---- Bibliography ----
% %
% % BibTeX users should specify bibliography style 'splncs04'.
% % References will then be sorted and formatted in the correct style.
% \bibliography{egbib}

\end{document}